%% file: cameral_ready.tex
\def\year{2022}\relax
\documentclass[letterpaper]{article} % DO NOT CHANGE THIS
\usepackage{aaai22}  % DO NOT CHANGE THIS
\usepackage{times}  % DO NOT CHANGE THIS
\usepackage{helvet}  % DO NOT CHANGE THIS
\usepackage{courier}  % DO NOT CHANGE THIS
\usepackage[hyphens]{url}  % DO NOT CHANGE THIS
\usepackage{graphicx} % DO NOT CHANGE THIS
\def\UrlFont{\rm}  % DO NOT CHANGE THIS
\usepackage{natbib}  % DO NOT CHANGE THIS AND DO NOT ADD ANY OPTIONS TO IT
\usepackage{caption} % DO NOT CHANGE THIS AND DO NOT ADD ANY OPTIONS TO IT
\DeclareCaptionStyle{ruled}{labelfont=normalfont,labelsep=colon,strut=off} % DO NOT CHANGE THIS
\usepackage{algorithm}
\usepackage{algorithmic}

\usepackage{mathtools}
\usepackage{amsfonts}
\usepackage{capt-of}
\usepackage{amsmath}
\usepackage{amsthm}
\usepackage{amssymb}
\usepackage{enumitem}

\usepackage{enumitem}
\usepackage{algorithm}
\usepackage{algorithmic}
\usepackage{wrapfig}
\usepackage{makecell}
\usepackage{booktabs}

\newtheorem{assumption}{Assumption}
\newtheorem{lemma}{Lemma}
\newtheorem{theorem}{Theorem}

\makeatletter

\newcommand{\Rmnum}[1]{\expandafter\@slowromancap\romannumeral #1@}
\makeatother

\usepackage{newfloat}
\usepackage{listings}
\floatstyle{ruled}
\newfloat{listing}{tb}{lst}{}
\floatname{listing}{Listing}
\title{Coordinating Momenta for Cross-silo Federated Learning}
\author {
    An Xu,
    Heng Huang\thanks{This work was supported by NSF IIS 1845666, 1852606, 1838627, 1837956, 1956002, IIA 2040588.} \\
}
\affiliations {
    Electrical and Computer Engineering Department, University of Pittsburgh, PA, USA \\
    an.xu@pitt.edu, heng.huang@pitt.edu
}

\begin{document}

\maketitle

\begin{abstract}
Communication efficiency is crucial for federated learning (FL). Conducting local training steps in clients to reduce the communication frequency between clients and the server is a common method to address this issue. However, this strategy leads to the client drift problem due to \textit{non-i.i.d.} data distributions in different clients which severely deteriorates the performance. In this work, we propose a new method to improve the training performance in cross-silo FL via maintaining double momentum buffers. In our algorithm, one momentum buffer is used to track the server model updating direction, and the other one is adopted to track the local model updating direction. More important, we introduce a novel momentum fusion technique to coordinate the server and local momentum buffers. We also derive the first theoretical convergence analysis involving both the server and local standard momentum SGD. Extensive deep FL experimental results verify that our new approach has a better training performance than the  FedAvg and existing standard momentum SGD variants.
\end{abstract}

\section{Introduction}\label{introduction}

With deep learning models becoming prevalent but data-hungry, data privacy emerges as an important issue. Federated learning (FL) \citep{konevcny2016federated} was thus introduced to leverage the massive data from different clients for training models without directly sharing nor aggregating data.
% In the beginning, the FL techniques focus on the \textit{cross-device} FL \cite{kairouz2019advances}, \emph{e.g.} the collaborative learning of mobile devices, which usually includes an extremely large number of clients. The well-known platforms include LEAF \cite{caldas2018leaf}, TensorFlow Federated \cite{tensorflowfl}, \emph{etc.}
More recently, an increasing number of FL techniques focus on addressing the cross-silo FL \cite{kairouz2019advances,gu2021privacy} problem, which has more and more real-world applications, such as the collaborative learning on health data across multiple medical centers \cite{liu2021feddg,guo2021multi} or financial data across different corporations and stakeholders.
% Several cross-silo FL platforms were recently released, \emph{e.g.} NVIDIA CLARA \cite{NVIDIA}, Intel OpenFL \cite{Intel}, \emph{etc.}
During the training, the server only communicates the model weights and updates with the participating clients.

However, the deep learning models require many training iterations to converge. Unlike workers in data-center distributed training with large network bandwidth and relatively low communication delay, the clients participating in the collaborative FL system are often faced with much more unstable conditions and slower network links due to the geo-distribution. Typically, \cite{bonawitz2019towards} showed that one communication round in FL could take about 2 to 3 minutes in practice. To address the communication inefficiency, the \textit{de facto} standard method FedAvg was proposed in \cite{mcmahan2017communication}. In FedAvg, the server sends clients the server model. Each client conducts many local training steps with its local data and sends back the updated model to the server. The server then averages the models received from clients and finishes one round of training. The increased local training steps can reduce the communication rounds and cost. Here we also refer to the idea of FedAvg as periodic averaging, which is closely related to local SGD \cite{stich2018local,yu2019linear}. Another parallel line of works is to compress the communication to reduce the volume of the message \cite{rothchild2020fetchsgd,karimireddy2019error,gao2021convergence,xu2021step,xu2020acceleration,xu2020optimal}. In this paper, we do not focus on communication compression.

Although periodic averaging methods such as FedAvg greatly improve the training efficiency in FL, a new problem named client drift arises. The data distributions of different clients are \textit{non-i.i.d.} because we cannot gather and randomly shuffle the client data as data-center distributed training does. Therefore, the stochastic gradient computed at different clients can be highly skewed. Given that we do many local training steps in each training round, skewed gradients will cause local model updating directions to gradually diverge and overfit local data at different clients. This client drift issue can deteriorate the performance of FedAvg drastically \cite{hsieh2020non,hsu2019measuring}, especially with a low similarity of the data distribution on different clients and a large number of local training steps.

As a method to reduce variance and smooth the model updating direction \cite{cutkosky2019momentum} to accelerate optimization, momentum SGD has shown its power in training many deep learning models in various tasks \cite{sutskever2013importance}. Local momentum SGD \cite{yu2019linear} (\emph{i.e.}, FedAvgLM) maintains momentum for the stochastic gradient in each training step but requires averaging local momentum buffer at the end of each training round. Therefore, FedAvgLM requires $\times 2$ communication cost compared with FedAvg. One strategy to achieve the same communication cost as FedAvg is resetting the local momentum buffer to zero at the end of each training round (FedAvgLM-Z) \cite{seide2016cntk,wang2018adaptive,wang2020tackling}. \citet{hsu2019measuring} and \citet{huo2020faster} proposed server momentum SGD (FedAvgSM) which maintains the momentum for the average local model update in a training round other than the stochastic gradient. The idea of FedAvgSM has been previously proposed in \cite{chen2016scalable} for speech models and in \cite{wang2019slowmo} for distributed training, but neither of them has applied it to FL. \citet{hsu2019measuring} and \citet{huo2020faster} empirically showed the ability of FedAvgSM to tackle client drift in FL, and \citet{wang2019slowmo} and \citet{huo2020faster} provided the convergence analysis of FedAvgSM. Throughout this paper, we refer to the naive combination of FedAvgSM and FedAvgLM(-Z) as FedAvgSLM(-Z). However, FedAvgSLM(-Z) has \textbf{no convergence guarantee} to the best of our knowledge. Moreover, there is a lack of understanding in \textbf{the connection between the server and local momenta} in FL. Whether we can further \textbf{improve standard momentum-based methods} in FL remains another question.

To address the above challenging problems, in this paper, we propose a new double momentum SGD (DOMO) algorithm, which leverages double momentum buffers to track the server and local model updating directions separately. We introduce a novel momentum fusion technique to coordinate the server and local momentum buffers. More importantly, we provide the theoretical analysis for the convergence of our new method for non-convex problems. Our new algorithm focuses on addressing the cross-silo FL problem, considering the recently increasing needs on it as described at the beginning of this section. We also regard it as time-consuming to compute the full local gradient and focuses on stochastic methods. We summarize our major technical contributions as follows:
\begin{itemize}[leftmargin=0.12in]
\setlength{\itemsep}{0pt}
    \item Propose a new double momentum SGD (DOMO) method with a novel momentum fusion technique.
    \item Derive the first convergence analysis involving both server and local standard momentum SGD in non-convex settings and under mild assumptions, show incorporating server momentum's convergence benefits over local momentum SGD for the first time, and provide new insights into their connection.
    \item Conduct deep FL experiments to show that DOMO can improve the test accuracy by up to 5\% compared with the state-of-the-art momentum-based method when training VGG-16 on CIFAR-10, while the naive combination FedAvgSLM(-Z) may sometimes hurt the performance compared with FedAvgSM.
\end{itemize}

\section{Background and Related Work}\label{related work}

FL can be formulated as an optimization problem of $\min_{\textbf{x}\in\mathbb{R}^d} \frac{1}{K}\sum^{K-1}_{k=0}f^{(k)}(\textbf{x})$, where $f^{(k)}$ is the local loss function on client $k$, $\textbf{x}$ is the model weights with $d$ as its dimension, and $K$ is the number of clients. Other basic notations are listed below. In FedAvg, the client trains the local model for $P$ steps using stochastic gradient $\nabla F^{(k)}(\textbf{x}^{(k)}_{r,p},\xi^{(k)}_{r,p})$ and sends local model update $\textbf{x}^{(k)}_{r,P}-\textbf{x}^{(k)}_{r,0}$ to server. Server then takes an average and updates the server model via $\textbf{x}_{r+1}=\textbf{x}_{r}-\frac{\alpha}{K}\sum^{K-1}_{k=0}(\textbf{x}^{(k)}_{r,P}-\textbf{x}^{(k)}_{r,0})$.

\textbf{Basic notations}:
\begin{itemize}[leftmargin=0.26in]
\setlength{\itemsep}{0pt}
    \item Training round (total): $r$ ($R$); Local training steps (total): $p$ ($P$); Client (total): $k$ ($K$);
    \item Global training step (total): $t=rP+p$ ($T=RP$); Server, local learning rate: $\alpha$, $\eta$;
    \item Momentum fusion constant $\beta$; Server, local momentum constant: $\mu_s$, $\mu_l$;
    \item Server, local momentum buffer: $\textbf{m}_r$, $\textbf{m}^{(k)}_{r,p}$; Server, (average) local model: $\textbf{x}_{r}$, $\textbf{x}^{(k)}_{r,p}$ ($\overline{\textbf{x}}^{(k)}_{r,p}$);
    \item Local stochastic gradient: $\nabla F^{(k)}(\textbf{x}^{(k)}_{r,p},\xi^{(k)}_{r,p})$, where $\xi^{(k)}_{r,p}$ is the sampling random variable;
    \item Local full gradient: $\nabla f^{(k)}(\textbf{x}^{(k)}_{r,p})=\mathbb{E}_{\xi^{(k)}_{r,p}}[\nabla F^{(k)}(\textbf{x}^{(k)}_{r,p},\xi^{(k)}_{r,p})]$ (unbiased sampling).
\end{itemize}

\textbf{Momentum-based.} State-of-the-art method server momentum SGD (FedAvgSM) maintains a server momentum buffer with the local model update $\frac{\alpha}{K}\sum^{K-1}_{k=0}(\textbf{x}^{(k)}_{r,P}-\textbf{x}^{(k)}_{r,0})$ to update the server model. While local momentum SGD maintains a local momentum buffer with $\nabla F^{(k)}(\textbf{x}^{(K)}_{r,p},\xi^{(k)}_{r,p})$ to update the local model. The communication costs compared with FedAvg are $\times 1$, $\times 2$ and $\times 1$ for FedAvgSM, FedAvgLM and FedAvgLM-Z respectively.

\textbf{Adaptive Methods.} \cite{reddi2020adaptive} applied the idea of using server statistics as in server momentum SGD to adaptive optimizers including Adam \cite{kingma2014adam}, AdaGrad \cite{duchi2011adaptive}, and Yogi \cite{zaheer2018adaptive}. \cite{reddi2020adaptive} showed that server learning rate should be smaller than $\mathcal{O}(1)$ in terms of complexity, but the exact value was unknown.

\textbf{Inter-client Variance Reduction.} Variance reduction in FL \cite{karimireddy2020scaffold,acar2020federated} refers to correct client drift caused by \textit{non-i.i.d.} data distribution on different clients following the variance reduction convention. In contrast, traditional stochastic variance reduced methods that are popular in convex optimization \cite{johnson2013accelerating,defazio2014saga} can be seen as intra-client variance reduction. Scaffold \cite{karimireddy2020scaffold} proposed to maintain a control variate $c_k$ on each client $k$ and add $\frac{1}{K}\sum^{K-1}_{k=0}c_k - c_k$ to gradient $\nabla F^{(k)}_{r,p}(\textbf{x}^{(k)}_{r,p},\xi^{(k)}_{r,p})$ when conducting local training. A prior work VRL-SGD \cite{liang2019variance} was built on a similar idea with $c_k$ equal to the average local gradients in the last training round. Both Scaffold and VRL-SGD have to maintain and communicate local statistics, which makes the clients stateful and requires $\times 2$ communication cost. Mime \cite{karimireddy2020mime} proposed to apply server statistics locally to address this issue. However, Mime has to compute the full local gradient which can be prohibitive in cross-silo FL. It also needs $\times 2$ communication cost. Besides, Mime's theoretical results are based on Storm \cite{cutkosky2019momentum} but their algorithm is based on Polyak's momentum. Though theoretically appealing, the variance reduction technique has shown to be ineffective in practical neural networks' optimization \cite{defazio2018ineffectiveness}. \cite{defazio2018ineffectiveness} showed that common tricks such as data augmentation, batch normalization \cite{ioffe2015batch}, and dropout \cite{srivastava2014dropout} broke the transformation locking and deviated practice from variance reduction's theory.

\textbf{Other.} There are some other settings of FL including heterogeneous optimization \cite{li2018federated,wang2020tackling}, fairness \cite{mohri2019agnostic,Li2020Fair,li2021ditto}, personalization \cite{t2020personalized,fallah2020personalized,jiang2019improving,shamsian2021personalized}, etc. These different settings, variance reduction techniques, and server statistics can sometimes be combined. Here we focus on momentum-based FL methods.

\section{New Double Momentum SGD (DOMO)}

\begin{algorithm}[t]
\caption{FL with double momenta.}
\label{framework alg}
\begin{algorithmic}[1]
    \STATE {\bfseries Input:} local training steps $P\geq 1$, \#rounds $R$, \#clients $K$, server (local) learning rate $\alpha$ ($\eta$), server (local) momentum constant $\mu_{s}$ ($\mu_l$), momentum fusion constant $\beta$.
    \STATE {\bfseries Initialize:} Server, local momentum buffer $\textbf{m}_0,\textbf{m}^{(k)}_{0,0}=\textbf{0}$. Local model $\textbf{x}^{(k)}_{0,0}=\textbf{x}_0$.
    
    \FOR{$r=0,1,\cdots,R-1$}
        \STATE \textbf{Client} $k$:
        \STATE ($r\geq 1$) Receive $\textbf{x}^{(k)}_{r,0}\leftarrow\textbf{x}_r$ . $\textbf{m}_{r}=\frac{1}{\alpha\eta P}(\textbf{x}_{r-1}-\textbf{x}_{r})$. $\textbf{m}^{(k)}_{r,0}\leftarrow \textbf{0}$. \hfill // \textit{Reset local momentum.}
        \FOR{$p=0,1,\cdots,P-1$}
            \STATE Option \Rmnum{1}: $\textbf{x}^{(k)}_{r,p}\leftarrow \textbf{x}^{(k)}_{r,p}-\eta\beta P \textbf{m}_{r}\cdot \textbf{1}_{p=0}$ // \textit{Pre-momentum fusion (DOMO)}
            \STATE $\textbf{m}^{(k)}_{r,p+1}=\mu_l\textbf{m}^{(k)}_{r,p} + \nabla F(\textbf{x}^{(k)}_{r,p},\xi^{(k)}_{r,p})$
            \STATE Option \Rmnum{1}: $\textbf{x}^{(k)}_{r,p+1}=\textbf{x}^{(k)}_{r,p}-\eta\textbf{m}^{(k)}_{r,p+1}$
            \STATE Option \Rmnum{2}: $\textbf{x}^{(k)}_{r,p+1}=\textbf{x}^{(k)}_{r,p}-\eta\textbf{m}^{(k)}_{r,p+1}-\eta\beta \textbf{m}_r$ // \textit{Intra-momentum fusion (DOMO-S)}
        \ENDFOR
        \STATE Send $\textbf{d}^{(k)}_{r}=\frac{1}{P}\sum^{P-1}_{p=0}\textbf{m}^{(k)}_{r,p+1}$ to the server.
        \STATE \textbf{Server:}
        \STATE Receive $\textbf{d}^{(k)}_{r}$ from client $k\in[K]$. $\textbf{m}_{r+1}=\mu_s\textbf{m}_{r}+\frac{1}{K}\sum^{K}_{k=1}\textbf{d}^{(k)}_{r}$.
        \STATE $\textbf{x}_{r+1}=\textbf{x}_{r}-\alpha\eta P\textbf{m}_{r+1}$. Send $\textbf{x}_{r+1}$ to client $k\in[K]$.
    \ENDFOR
    
    \STATE {\bfseries Output: $\textbf{x}_{R}$} 
\end{algorithmic}
\end{algorithm}

In this section, we address the connection and coordination of server and local momenta to improve momentum-based methods by introducing our new double momentum SGD (DOMO) algorithm. We maintain both the server and local statistics (momentum buffers). Nevertheless, the local momentum buffer does not make the clients stateful because the local momentum buffer will be reset to zero at the end of each training round for every client.

\textbf{Motivation.} We observe that FedAvgSM applies momentum SGD update after aggregating the local model update from clients in the training round $r$. Specifically, it updates the model at the server via $\textbf{x}_{r+1}=\textbf{x}_r-\alpha\eta P\textbf{m}_{r+1}=\textbf{x}_r-\alpha\eta\mu_s\textbf{m}_r-\frac{\alpha\eta P}{K}\sum^{K}_{k=1}\textbf{d}^{(k)}_r$ (Algorithm \ref{framework alg} lines 14 and 15). We can see that the server momentum buffer $\textbf{m}_r$ is only applied at the \textit{server} side after the clients finish the training round $r$. Therefore, FedAvgSM fails to take advantage of the server momentum buffer $\textbf{m}_r$ during the local training at the \textit{client} side in the training round $r$. The same issue exists for FedAvgSLM(-Z), where the local optimizer is also momentum SGD.

Recognizing this issue, we propose DOMO (summarized in Algorithm \ref{framework alg} where \textbf{1} denotes the indicator function) by utilizing server momentum statistics $\textbf{m}_r$ to help local training in round $r$ at the \textit{client} side as it provides information on global updating direction. The whole framework can be briefly summarized in the following steps.

\begin{enumerate}[leftmargin=0.2in]
\setlength{\itemsep}{0pt}
\item Receive the initial model from the server at the beginning of the training round.
\item Fuse the server momentum buffer in local training steps.
\item Remove the server momentum's effect from the local model update before sending it to the server.
\item Aggregate local model updates from clients to update model and statistics at the server.
\end{enumerate}

To avoid incurring additional communication cost than FedAvg, we 1) infer the server momentum buffer $\textbf{m}_r$ via the current and last initial model ($\textbf{x}_{r},\textbf{x}_{r-1}$), and 2) reset the local momentum buffer $\textbf{m}^{(k)}_{r,0}$ to zero instead of averaging. Note that for FedAvgSLM, the local momentum buffer has to be averaged.

\textbf{Momentum Fusion in Local Training Steps.} In each local training step, the local momentum buffer is updated following the standard momentum SGD method (Algorithm \ref{framework alg} line 8). We propose two options to fuse server momentum into local training steps. The default Option \Rmnum{1} is DOMO and the Option \Rmnum{2} is DOMO-S with ``S'' standing for ``scatter''. In DOMO, we apply server momentum buffer $\textbf{m}_r$ with coefficient $\beta P$ and learning rate $\eta$ to the local model before the local training starts. $\beta$ is the momentum fusion constant. DOMO-S is an heuristic extension of DOMO by evenly scattering this procedure to all the $P$ local training steps. Therefore, the coefficient becomes $\beta$ instead of $\beta P$. Intuitively, the local model updating direction should be adjusted by the direction of the server momentum buffer to alleviate the client drift issue. Furthermore, DOMO-S follows this motivation in a more fine-grained way by adjusting each local momentum SGD training step with the server momentum buffer.

\textbf{Pre-Momentum, Intra-Momentum, Post-Momentum.} To help understand the connection between the server and local momenta better, here we propose new concepts called pre-momentum, intra-momentum, and post-momentum. The naive combination FedAvgSLM(-Z) can be interpreted as post-momentum because the current server momentum buffer $\textbf{m}_r$ is applied at the end of the training round and after all the local momentum SGD training steps are finished. Therefore, FedAvgSLM(-Z) has no momentum fusion to help local training. While our proposed DOMO can be regarded as pre-momentum because it applies the current server momentum buffer $\textbf{m}_r$ at the beginning of the training round ($p=0$) and before the local momentum SGD training starts. Similarly, the proposed DOMO-S works as intra-momentum because it scatters the effect of the current server momentum buffer $\textbf{m}_r$ to the whole local momentum SGD training steps. These new concepts shed new insights for the connection and coordination between server and local momenta by looking at the order of applying server momentum buffer and local momentum buffer. Considering that server and local momentum buffers can be regarded as the smoothed server and local model updating direction, the order of applying which one first should not make much difference when the similarity of data distribution across clients is high, \textit{i.e.}, the client drift issue is not severe. However, when the data similarity is low, it becomes more critical to provide the information of server model updating direction during the local training as DOMO (pre-momentum) and DOMO-S (intra-momentum) do.

\textbf{Aggregate Local Model Updates without Server Momentum.} We propose to remove the effect of server momentum $\textbf{m}_{r}$ in local model updates (Algorithm \ref{framework alg} line 13) when aggregating them to server. The equivalent server momentum constant would have been deviated to $\mu_s + \beta$ if we would not remove it.

\section{Convergence Analysis}\label{convergence}

\begin{figure*}[t]
    \centering
    \includegraphics[width=.315\textwidth]{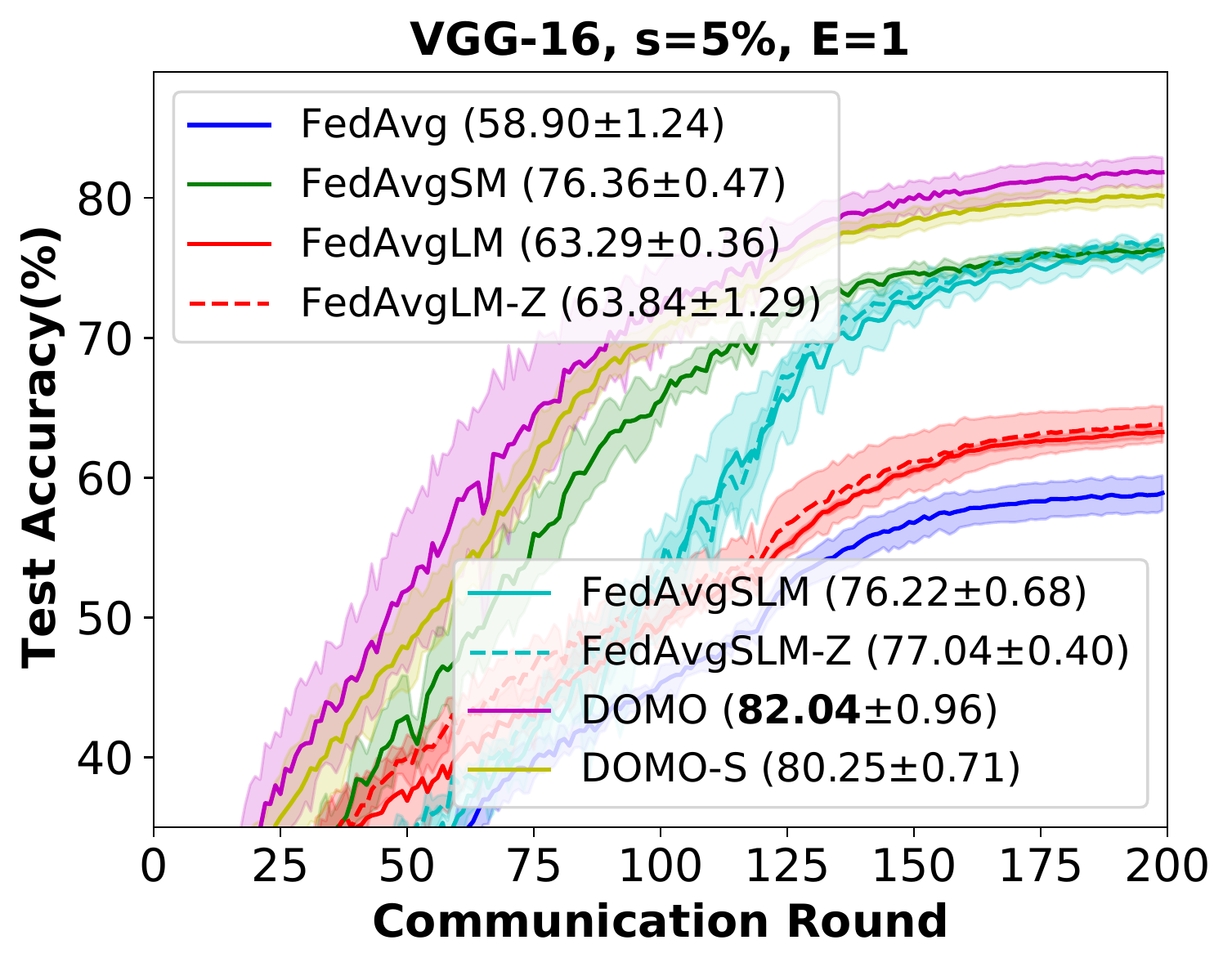}
    \includegraphics[width=.315\textwidth]{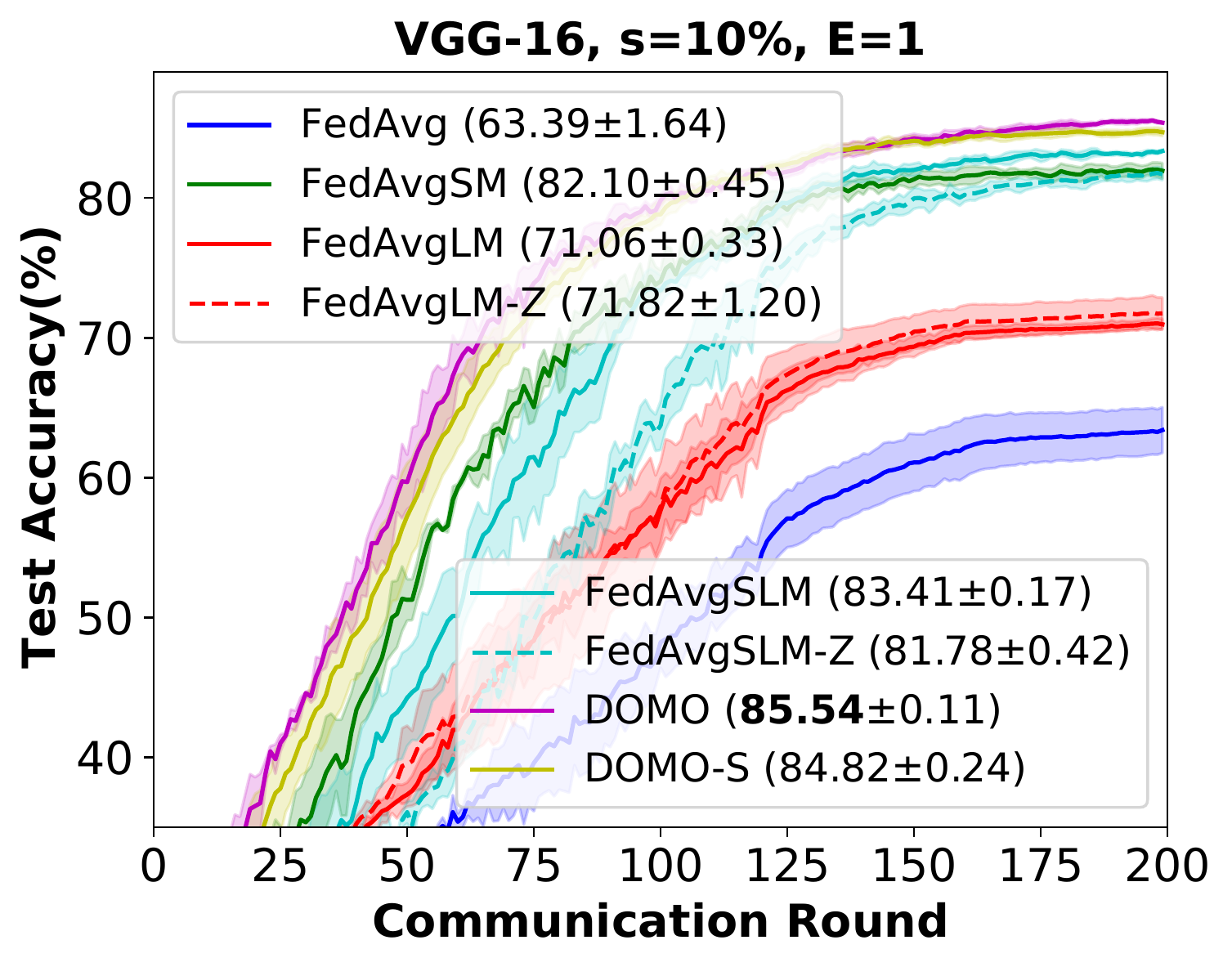}
    \includegraphics[width=.315\textwidth]{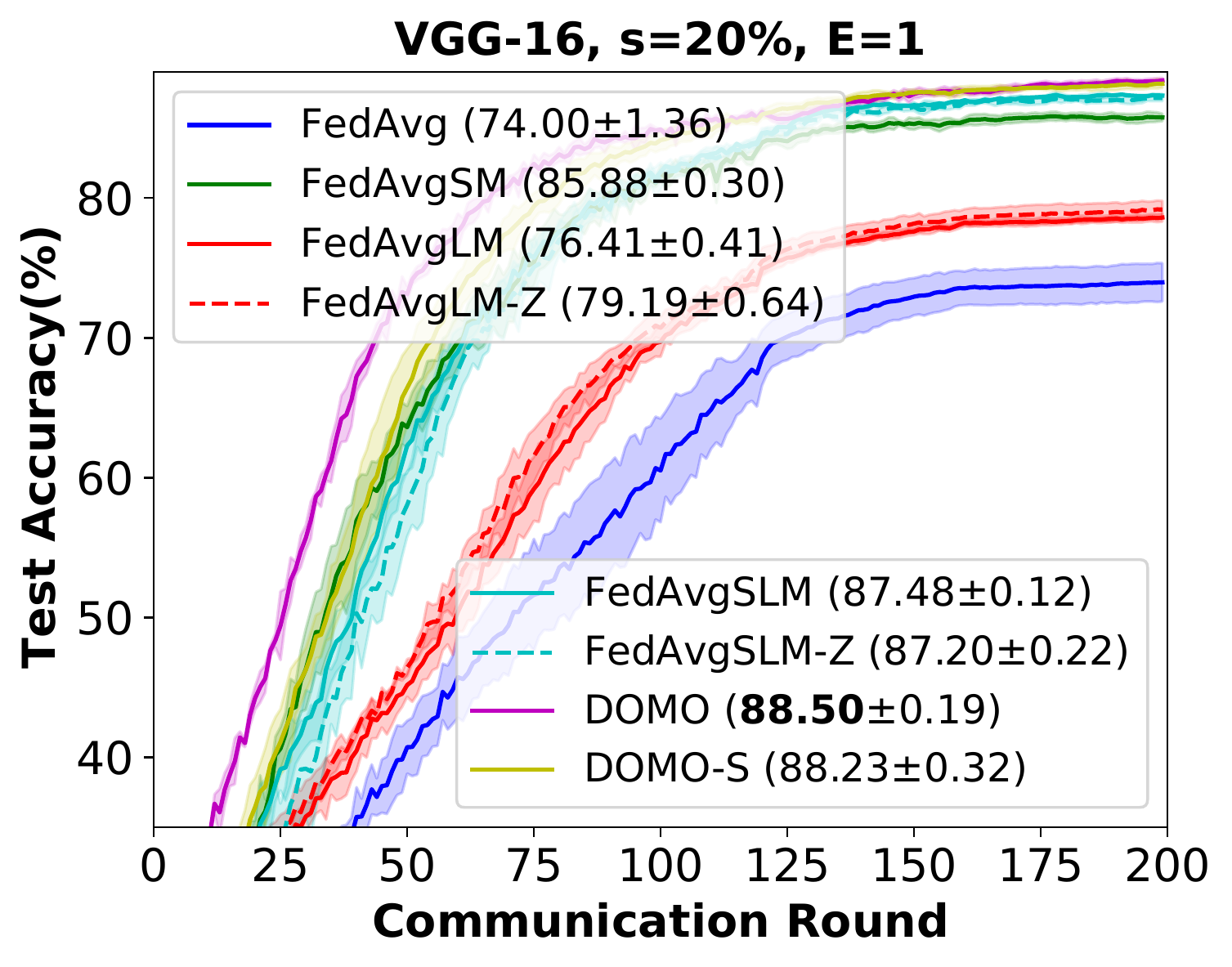}
    \caption{CIFAR-10 training curves using the VGG-16 model with various data similarity $s$.}
    \label{fig:training curves vary s}
\end{figure*}

\begin{figure*}[t]
    \centering
    \includegraphics[width=.315\textwidth]{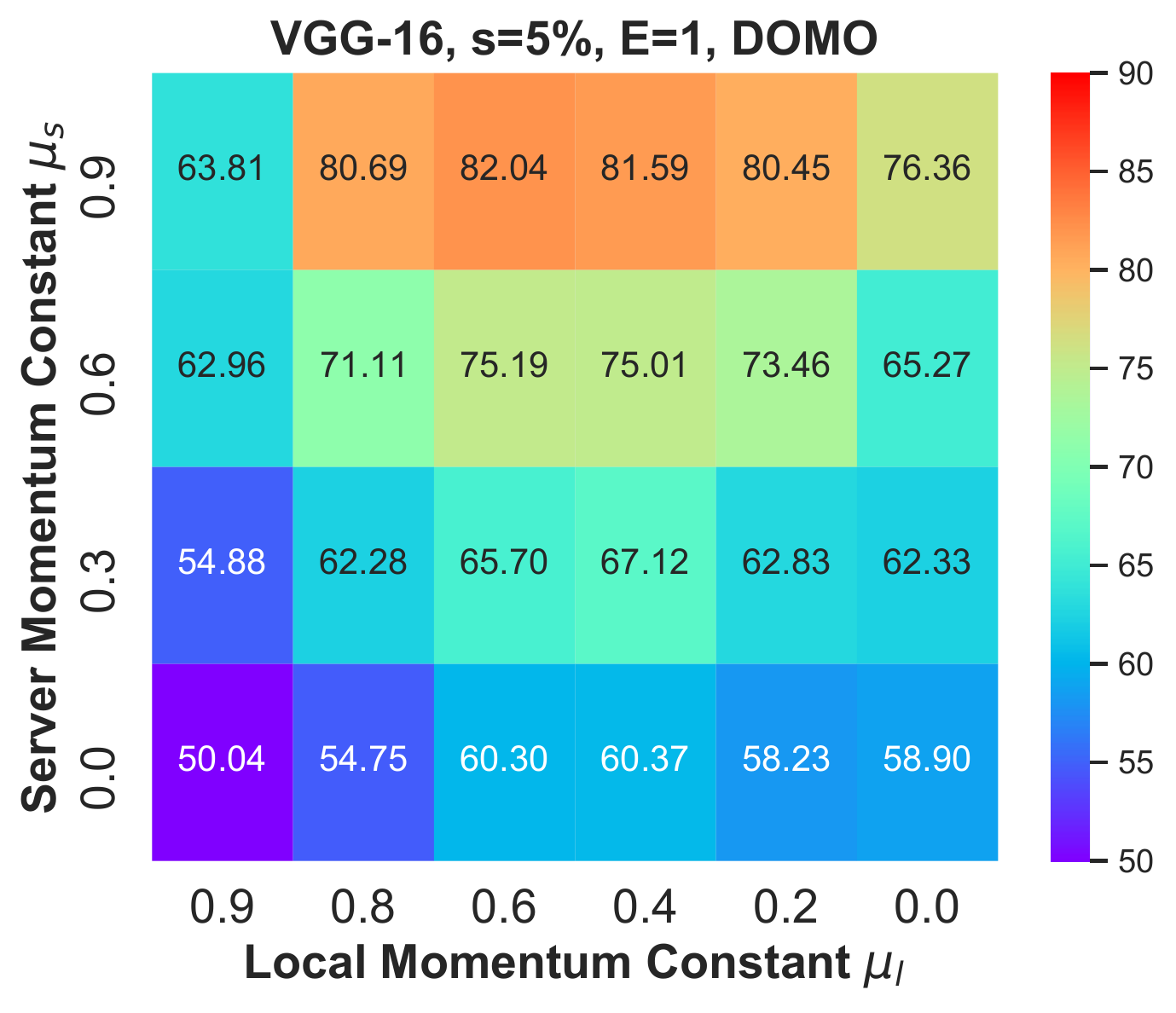}
    \includegraphics[width=.315\textwidth]{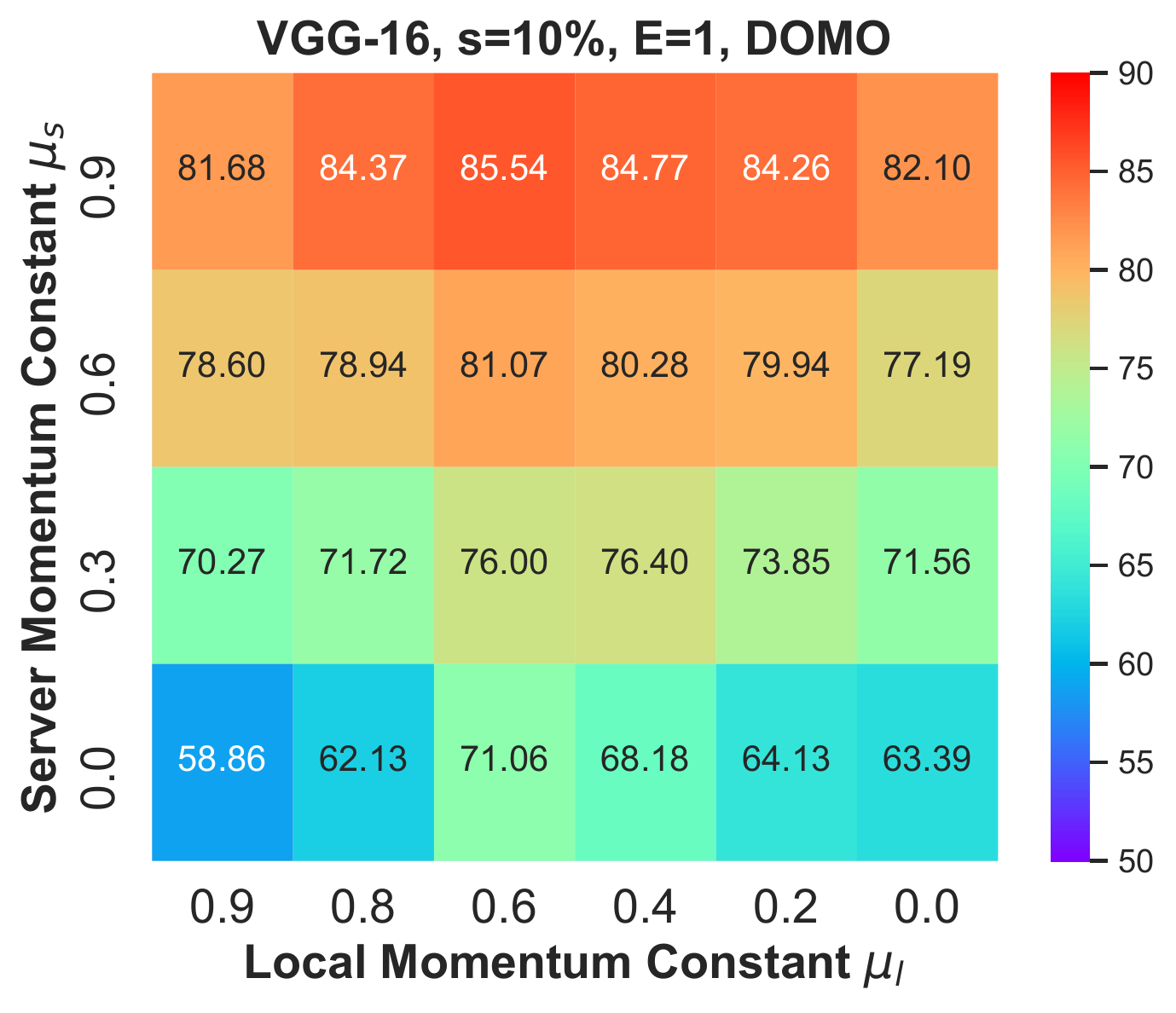}
    \includegraphics[width=.315\textwidth]{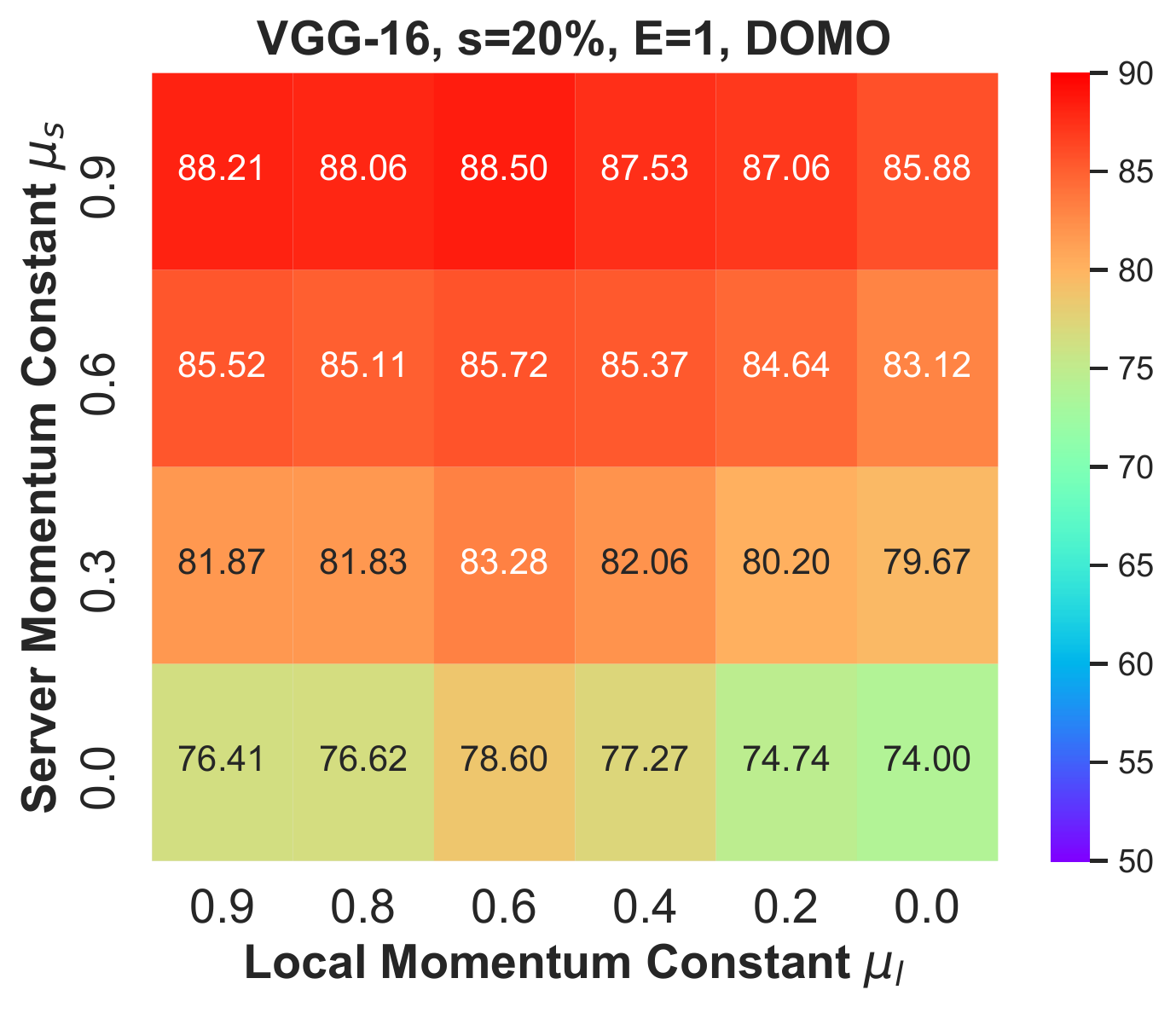}
    \caption{CIFAR-10 test accuracy (\%) with various sever momentum constant $\mu_s$ and local momentum constant $\mu_l$. $\mu_s=0$ corresponds to FedAvgLM, $\mu_l=0$ corresponds to FedAvgLM, $\mu_s=0\,\&\,\mu_l=0$ corresponds to FedAvg, and $\mu_s\neq 0\,\&\,\mu_l \neq 0$ corresponds to DOMO.}
    \label{fig:heatmap}
\end{figure*}

\begin{figure*}[t]
    \centering
    \includegraphics[width=.315\textwidth]{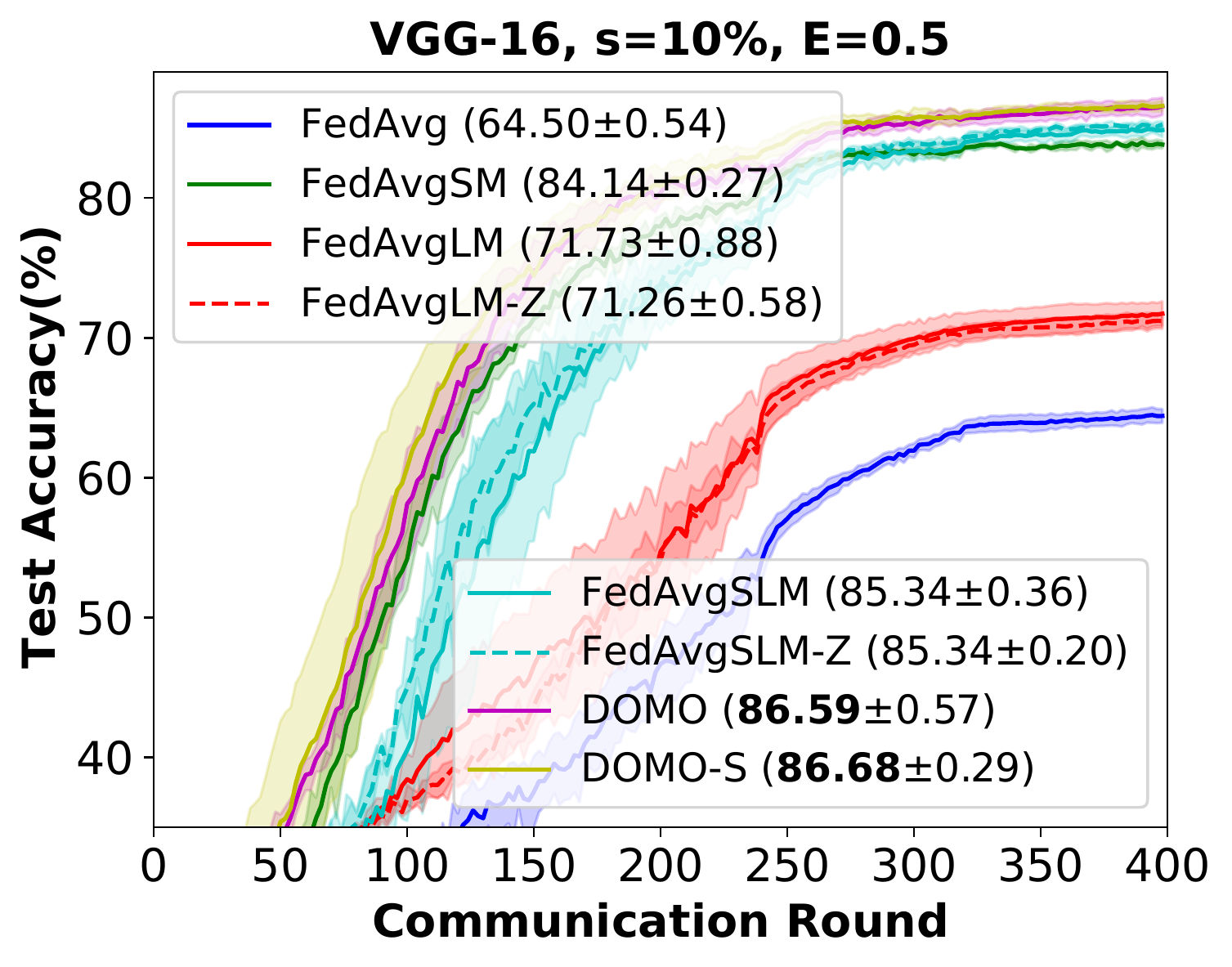}
    \includegraphics[width=.315\textwidth]{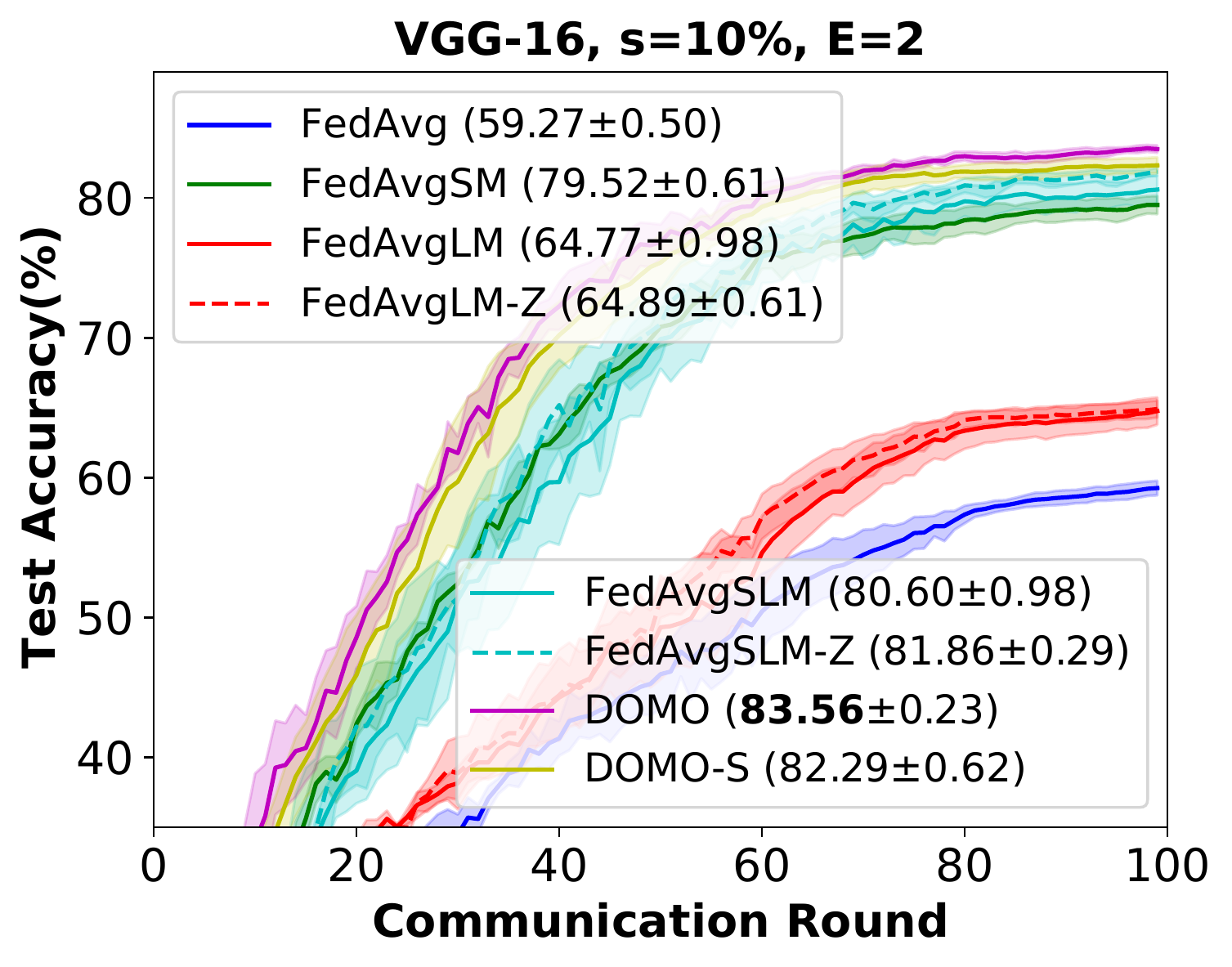}
    \includegraphics[width=.315\textwidth]{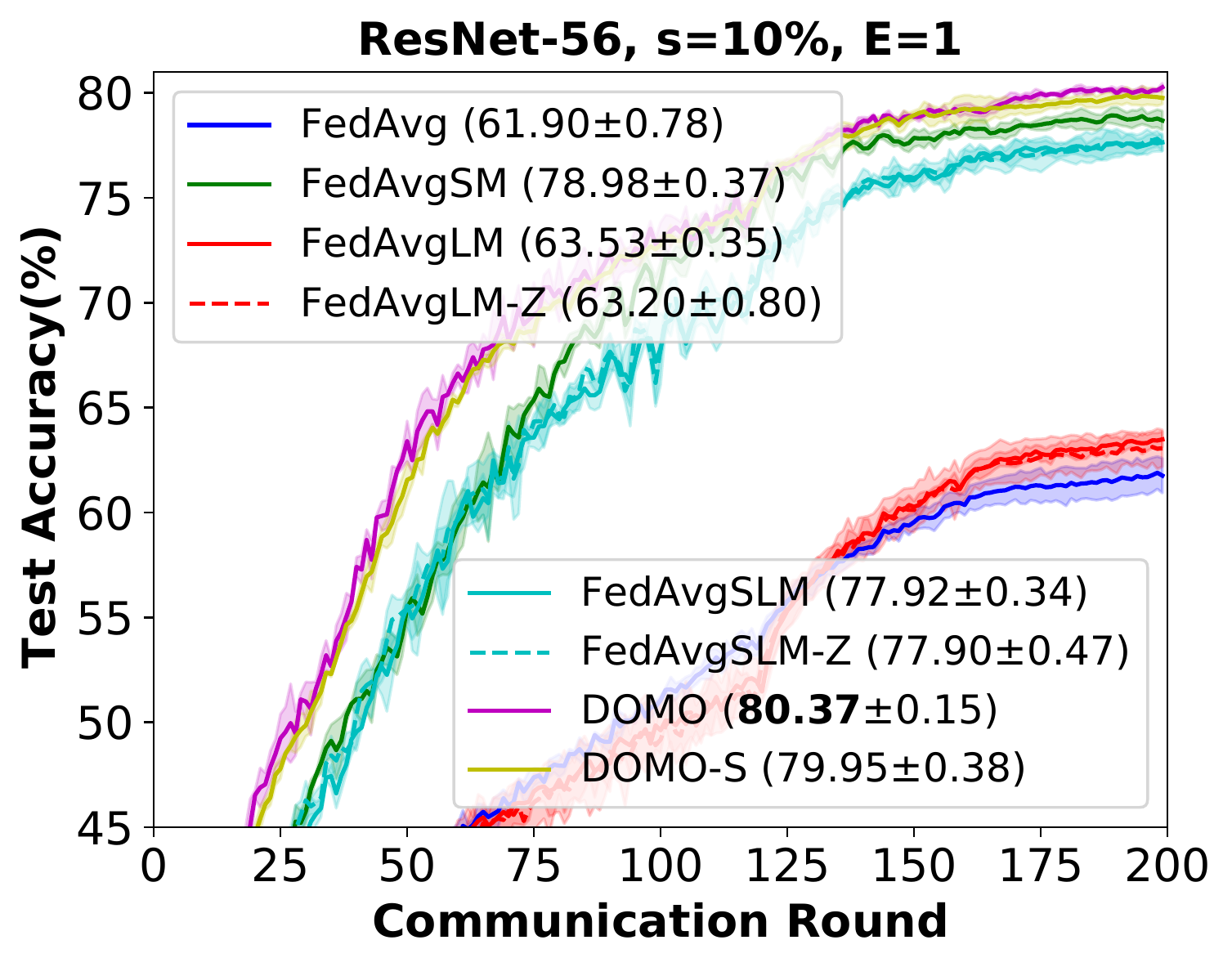}
    \caption{Left and Middle: CIFAR-10 training curves using the VGG-16 model with various local epoch $E$. $E=1$ has been shown in the middle plot of Figure \ref{fig:training curves vary s} and is not repeatedly shown here. Right: CIFAR-10 training curves using the ResNet-56 model.}
    \label{fig:training curves vary e}
\end{figure*}
\begin{table*}[t]
\small
    \caption{CIFAR-10 test accuracy (\%) when training VGG-16 using DOMO with various hyper-parameters $\alpha$ and $\beta$. Data similarity $s=10\%$ and local epoch $E=1$. $\alpha$ is fixed at 1.0 with various $\beta$ in the first column, while $\beta$ is fixed at 0.9 with various $\alpha$ in the second column.}
    \label{tab:vary alpha beta}
    \centering
    \begin{tabular}{cc|cc|cc|cc}
        \toprule
        $\beta$ & $\alpha=1.0$ & $\beta$ & $\alpha=1.0$ & $\alpha$ & $\beta=0.9$ & $\alpha$ & $\beta=0.9$ \\
        \midrule
        1.0 & 81.89 $\pm$ 0.40 & 0.6 & 81.60 $\pm$ 0.40 & 1.0 & \textbf{85.54} $\pm$ 0.11 & 0.6 & 83.76 $\pm$ 0.28 \\
        0.9 & \textbf{85.54} $\pm$ 0.11 & 0.4 & 77.80 $\pm$ 0.88 & 0.9 & 84.63 $\pm$ 0.64 & 0.4 & 82.08 $\pm$ 0.50 \\
        0.8 & 83.84 $\pm$ 0.56 & 0.2 & 74.54 $\pm$ 0.49 & 0.8 & 84.83 $\pm$ 0.56 & 0.2 & 77.58 $\pm$ 0.62 \\
        \bottomrule
    \end{tabular}
\end{table*}

\begin{table*}[t]
\small
    \caption{SVHN test accuracy (\%) when training ResNet-20.}
    \label{tab:svhn}
    \centering
    \begin{tabular}{cccccc}
        \toprule
        FedAvg & FedAvgSM & FedAvgLM(-Z) &  FedAvgSLM(-Z) & DOMO-S & DOMO \\
        \midrule
        \makecell{87.79 $\pm$ 0.72} & \makecell{88.81 $\pm$ 0.49} & \makecell{88.86 $\pm$ 0.19 \\ (87.93 $\pm$ 0.98)}  & \makecell{88.67 $\pm$ 0.32 \\ (88.89 $\pm$ 0.62)} & \textbf{90.45} $\pm$ 0.56 & 90.34 $\pm$ 0.83\\
        \bottomrule
    \end{tabular}
\end{table*}

\begin{table*}[t]
\small
    \caption{CIFAR-100 test accuracy (\%). Second row: VGG-16. Third row: ResNet-56.}
    \label{tab:cifar100}
    \centering
    \begin{tabular}{cccccccc}
        \toprule
        FedAvg & FedAvgSM & FedAvgLM(-Z) & FedAvgSLM(-Z) & DOMO-S & DOMO \\
        \midrule
        \makecell{20.77 $\pm$ 1.31} & \makecell{35.14 $\pm$ 1.70} & \makecell{38.29 $\pm$ 1.01 \\ (35.45 $\pm$ 2.04)} & \makecell{60.01 $\pm$ 0.28 \\ (57.89 $\pm$ 2.79)} & 61.69 $\pm$ 0.41 & \textbf{62.47} $\pm$ 0.73 \\
        \midrule
        \makecell{39.61 $\pm$ 0.66} & \makecell{61.92 $\pm$ 0.43} & \makecell{46.65 $\pm$ 1.38 \\ (45.09 $\pm$ 0.26)} & \makecell{62.95 $\pm$ 0.51 \\ (63.45 $\pm$ 0.61)} & 64.34 $\pm$ 0.59 & \textbf{65.84} $\pm$ 0.30\\
        \bottomrule
    \end{tabular}
\end{table*}

In this section, we will discuss our convergence analysis framework with double momenta and the potential difficulty for the naive combination FedAvgSLM(-Z). There has been little theoretical analysis in existing literature for FedAvgSLM(-Z) possibly due to this theoretical difficulty. After that, we will show how the motivation of DOMO addresses it. This is the first convergence analysis involving both server and local standard momentum SGD to the best of our knowledge. Both resetting and averaging local momentum buffer are considered, though we only reset it in Algorithm \ref{framework alg} for less communication.  Please refer to the Supplementary Material for proof details.

We consider non-convex smooth objective function satisfying Assumption \ref{Lipschitz Smoothness}. We also assume that the local stochastic gradient is an unbiased estimation of the local full gradient and has a bounded variance in Assumption \ref{Bounded Variance}. Furthermore, we bound the \textit{non-i.i.d.} data distribution across clients in Assumption \ref{bounded noniid}, which is widely employed in existing works such as \cite{yu2019linear,reddi2020adaptive,wang2019slowmo,karimireddy2020mime}. $G$ measures the data similarity in different clients and $G=0$ corresponds to \textit{i.i.d.} data distribution as in data-center distributed training. A low data similarity will lead to a larger $G^2$. For simplicity, let $f_*$ denote the optimal global objective value. Other basic notations have been summarized in Section ``Background \& Related Works".
\begin{assumption}\label{Lipschitz Smoothness}
($L$-Lipschitz Smoothness) The global objective function $f(\cdot)$ and local objective function $f^{(k)}$ are $L$-smooth, i.e., $\|\nabla f^{(k)}(\textbf{x})-\nabla f^{(k)}(\textbf{y})\|_2 \leq L\|\textbf{x}-\textbf{y}\|_2$ and $\|\nabla f(\textbf{x})-\nabla f(\textbf{y})\|_2 \leq L\|\textbf{x}-\textbf{y}\|_2, \forall \textbf{x}, \textbf{y} \in \mathbb{R}^d, k\in[K]$.
\end{assumption}
\begin{assumption}\label{Bounded Variance}
(Unbiased Gradient and Bounded Variance) The stochastic gradient $\nabla F^{(k)}(\textbf{x},\xi)$ is an unbiased estimation of the full gradient $\nabla f^{(k)}(\textbf{x})$, i.e., $\mathbb{E}_{\xi}\nabla F(\textbf{x},\xi)=\nabla f(\textbf{x}), \forall \textbf{x}\in\mathbb{R}^d$. Its variance is also bounded, \textit{i.e.}, $\mathbb{E}_{\xi}\|\nabla F(\textbf{x},\xi)-\nabla f(\textbf{x})\|^2_2\leq \sigma^2, \forall \textbf{x}\in\mathbb{R}^d$.
\end{assumption}
\begin{assumption}\label{bounded noniid}
(Bounded \textit{Non-i.i.d.} Distribution \cite{yu2019linear,reddi2020adaptive,wang2019slowmo,karimireddy2020mime}) For any client $k\in[K]$ and  $\textbf{x} \in \mathbb{R}^d$, there exists $B \geq 0$ and $G \geq 0$, the variance of the local full gradient in each client is upper bounded so that $\frac{1}{K}\sum^{K-1}_{k=0}\| \nabla f^{(k)}(\textbf{x}) - \nabla f(\textbf{x})\|_2^2 \leq G^2$.
\end{assumption}

\begin{lemma}\label{auxiliary variable}
(DOMO updating rule) Let $0\leq r \leq R-1$ and $0\leq p\leq P-1$. Let $\widehat{\textbf{y}}_{r,p}=\textbf{x}_0-\frac{\alpha\eta}{(1-\mu_s)K}\sum^{K-1}_{k=0}\sum^{r}_{r^\prime=0}\sum^{p-1}_{p^\prime=0}\textbf{m}^{(k)}_{r^\prime,p^\prime+1}$ and $\textbf{z}_{r,p}=\frac{1}{1-\mu_l}\widehat{\textbf{y}}_{r,p}-\frac{\mu_l}{1-\mu_l}\widehat{\textbf{y}}_{r,p-1}$ where $\widehat{\textbf{y}}_{0,-1}=\widehat{\textbf{y}}_{0,0}=\textbf{x}_0$, then
\begin{equation*}
    \textbf{z}_{r,p+1}=\textbf{z}_{r,p} - \frac{\alpha\eta}{(1-\mu_l)(1-\mu_s)K}\sum^{K-1}_{k=0}\nabla F^{(k)}(\textbf{x}^{(k)}_{r,p},\xi^{(k)}_{r,p})
\end{equation*}
\end{lemma}

The key of the proof is to find a novel auxiliary sequence $\{\textbf{z}_{r,p}\}$ that not only has a concise update rule than the mixture of server and local momentum SGD, but also is close to the average local model $\{\overline{\textbf{x}}_{r,p}\}$. One difficulty is to analyze the server model update at the end of the training round ($\overline{\textbf{x}}_{r,P}\rightarrow \textbf{x}_{r+1}$) due to server momentum. To tackle it, we design $\textbf{z}_{r,P}=\textbf{z}_{r+1,0}$ to facilitate the analysis at the end of the training round. Lemma \ref{auxiliary variable} gives such an auxiliary sequence. Before to analyze the convergence of $\{\overline{\textbf{x}}_{r,p}\}$ with the help of $\{\textbf{z}_{r,p}\}$, we only have to bound $\|\textbf{z}_{r,p}-\overline{\textbf{x}}_{r,p}\|^2_2$ (\textbf{inconsistency bound}) and $\frac{1}{K}\sum^{K-1}_{k=0}\|\overline{\textbf{x}}_{r,p}-\textbf{x}^{(k)}_{r,p}\|^2_2$ (\textbf{divergence bound}). The divergence bound measures how the local models on different clients diverges and is more straightforward to analyze since it is only affected by local momentum SGD. The inconsistency bound measures the inconsistency between the auxiliary variable and the average local model as a trade-off for a more concise update rule.

\begin{lemma}\label{inconsistency bound}
(Inconsistency Bound) For DOMO, we have $(\textbf{z}_{r,p}-\overline{\textbf{x}}_{r,p})_{\text{DOMO}}=(1-\frac{\alpha}{1-\mu_s})\frac{\eta}{K}\sum^{K-1}_{k=0}\sum^{p-1}_{p^\prime=0}\textbf{m}^{(k)}_{r,p^\prime+1} - \frac{\mu_l\eta}{(1-\mu_l)K}\sum^{K-1}_{k=0}\textbf{m}^{(k)}_{r,p}$; while for FedAvgSLM(-Z), we have $(\textbf{z}_{r,p}-\overline{\textbf{x}}_{r,p})_{\text{FedAvgSLM(-Z)}}=(\textbf{z}_{r,p}-\overline{\textbf{x}}_{r,p})_{\text{DOMO}} + \frac{\mu_s}{1-\mu_s}\alpha\eta P\textbf{m}_{r}$.

Furthermore, assume that $\alpha\geq (1-\mu_s)(1-\mu_l)$, let $h=\frac{\alpha}{1-\mu_s}\frac{1+\mu_l-\mu_l^p}{1-\mu_l}-1$ for DOMO and $h=\frac{\mu_l}{1-\mu_l}$ for FedAvgLM(-Z), and we have
\begin{equation}\label{inconsistency bound eq}
\begin{split}
    &\sum^{R-1}_{r=0}\sum^{P-1}_{p=0}\|\textbf{z}_{r,p}-\overline{\textbf{x}}_{r,p}\|^2_2\leq \frac{\eta^2}{1-\mu_l}(\sum^{P-1}_{p=0}\frac{h^2\mu_l^p}{1-\mu_l^P}) \cdot\\
    &\sum^{R-1}_{r=0}\sum^{P-1}_{p=0}\|\frac{1}{K}\sum^{K-1}_{k=0}\nabla F^{(k)}(\textbf{x}^{(k)}_{r,p},\xi^{(k)}_{r,p})\|^2_2\,.
\end{split}
\end{equation}
\end{lemma}

\textbf{Theoretical Difficulty for FedAvgSLM(-Z) but Addressed by DOMO.} From Lemma \ref{inconsistency bound}, we can see that without momentum fusion, the inconsistency bound for FedAvgSLM-Z has an additional term related to $P\textbf{m}_{r}$ compared with DOMO. For the corresponding inconsistency bound, this term will lead to $\sum^{R-1}_{r=0}\sum^{P-1}_{p=0}\|P\textbf{m}_{r}\|^2_2=P^2\sum^{R-1}_{r=0}\sum^{P-1}_{p=0}\|\textbf{m}_{r}\|^2_2$. Intuitively, if we ignore the constant and simply assume that $\|\textbf{m}_{r}\|^2_2$ is of the same complexity as $\|\nabla F\|^2_2$, then it causes an additional term of complexity $\mathcal{O}(RP^3\|\nabla F\|^2_2)$ in the inconsistency bound, much larger than the complexity $\mathcal{O}(RP^2\|\nabla F\|^2_2)$ for DOMO in the R.H.S. of Eq. (\ref{inconsistency bound eq}). This also means that FedAvgSLM(-Z) is more sensitive to $P$ and may even hurt the performance when $P$ is large as in FL.

\textbf{Tighten Inconsistency Bound with DOMO.} In Lemma \ref{inconsistency bound}, we also show that DOMO can tighten the inconsistency bound. Specifically, set $\alpha=(1-\mu_s)(1-\mu_l)$ and DOMO can scale the inconsistency bound down to about $(1-\mu_l)^2$ of that in FedAvgLM(-Z) ($\alpha=1, \mu_s=0$). Take the popular momentum constant $\mu_l=0.9$ as an example, the inconsistency bound of DOMO is reduced to $(1-\mu_l)^2=1\%$ compared with FedAvgLM(-Z). Therefore, momentum fusion not only addresses the difficulty for FedAvgSLM(-Z), but also helps the local momentum SGD training. To the best of our knowledge, this is the first time to show that \textit{incorporating server momentum leads to convergence benefits over local momentum}. It is also intuitively reasonable in that server momentum buffer carries historical local momentum information. But we note that this improvement analysis has not reached the optimal yet due to inequality scaling. Therefore, we fine-tune $\alpha$ and $\beta$ for the best performance in practice.

When $\alpha=1-\mu_s$ and $\beta=\mu_s$, we can see that DOMO has the same inconsistency bound as FedAvgLM(-Z) ($\alpha=1,\mu_s=0$). Consider the momentum buffer as a smoothed updating direction. Suppose the update of server momentum buffer $\textbf{m}_{r+1}=\mu_s\textbf{m}_r+\Delta_r$ becomes steady with $\Delta_r\rightarrow \Delta$ (the sum of local momentum buffer in local training), then $\textbf{m}_{r}$ will be approximately equal to $\frac{\Delta}{1-\mu_s}$. The coefficient $\frac{1}{1-\mu_s}$ leads to a different magnitude of the server and local momenta. Setting $\beta=\mu_s$ in Lemma \ref{inconsistency bound} gives $\alpha=1-\mu_s$, balancing the difference by a smaller server learning rate.

With the above lemmas, we have the following convergence analysis theorem for our new algorithm.
\begin{theorem}\label{convergence rate}
(Convergence of DOMO) Assume Assumptions \ref{Lipschitz Smoothness}, \ref{Bounded Variance} and \ref{bounded noniid} exist. Let $P\leq\frac{1-\mu_l}{6\eta L}$, $1-2\eta L-\frac{4\mu_l^2\eta^2L^2}{(1-\mu_l)^4}\geq 0$, $\alpha=1-\mu_s$ and $\beta=\mu_s$. For DOMO with either resetting or averaging local momentum buffer in Algorithm \ref{framework alg} line 5, we have
\begin{align}\label{convergence rate eq}
    &\frac{1}{RP}\sum^{R-1}_{r=0}\sum^{P-1}_{p=0}\mathbb{E}\|\nabla f(\overline{\textbf{x}}_{r,p})\|^2_2 \leq \frac{2(1-\mu_l)(f(\textbf{x}_0)-f_*)}{\eta RP} +\notag\\
    &\frac{9\eta^2L^2P^2G^2}{(1-\mu_l)^2}\notag + \frac{\eta L\sigma^2}{(1-\mu_l)}(\frac{1}{K}+\frac{3\eta LP}{2(1-\mu_l)}+\frac{2\mu_l^2\eta L}{(1-\mu_l)^4K})
\end{align}
\end{theorem}
\textbf{Complexity.} According to Theorem \ref{convergence rate}, let $\eta=\mathcal{O}(K^{\frac{1}{2}}R^{-\frac{1}{2}}P^{-\frac{1}{2}})$ and $P=\mathcal{O}(K^{-1}R^{\frac{1}{3}})$, then we have a convergence rate $\frac{1}{RP}\sum^{R-1}_{r=0}\sum^{P-1}_{p=0}\mathbb{E}\|\nabla f(\overline{\textbf{x}}_{r,p})\|^2_2=\mathcal{O}(K^{-\frac{1}{2}}R^{-\frac{1}{2}}P^{-\frac{1}{2}})$ regarding iteration complexity, which achieves a linear speedup regarding the number of clients $K$. DOMO also has a communication complexity of $\frac{1}{P}$ when resetting local momentum buffer, which increases with a larger number of clients $K$ but decreases with a larger communication rounds $R$. It becomes $\frac{2}{P}$ for averaging local momentum buffer, but $2$ is a constant and does not affect the theoretical complexity. Note that there is no $\mu_s$ in Theorem \ref{convergence rate} because it is eliminated in Lemma \ref{inconsistency bound} by setting $\beta=\mu_s$.

\section{Experimental Results}

\subsection{Settings}
All experiments are implemented using PyTorch \cite{paszke2019pytorch} and run on a cluster where each node is equipped with 4 Tesla P40 GPUs and 64 Intel(R) Xeon(R) CPU E5-2683 v4 cores @ 2.10GHz. We compare the following momentum-based FL methods: 1) FedAvg, 2) FedAvgSM (\textit{i.e.}, server momentum SGD), 3) FedAvgLM (\textit{i.e.}, local momentum SGD), 4) FedAvgLM-Z (\textit{i.e.}, local momentum SGD with resetting local momentum buffer), 5) FedAvgSLM (\textit{i.e.}, FedAvgSM + FedAvgLM), 6) FedAvgSLM-Z (\textit{i.e.}, FedAvgSM + FedAvgLM-Z, Algorithm \ref{framework alg} Option \Rmnum{3}), 7) DOMO (\textit{i.e.}, Algorithm \ref{framework alg} Option \Rmnum{1}), and 8) DOMO-S (\textit{i.e.}, Algorithm \ref{framework alg} Option \Rmnum{2}). In particular, FedAvg, FedAvgSM, FedAvgLM-Z, FedAvgSLM-Z, DOMO and DOMO-S have the same communication cost. FedAvgLM and FedAvgSLM need $\times 2$ communication cost.

We perform careful hyper-parameters tuning for all methods. The local momentum constant $\mu_l$ is selected from \{0.9, 0.8, 0.6, 0.4, 0.2\}. We select the server momentum constant $\mu_s$ from \{0.9, 0.6, 0.3\}. The base learning rate is selected from \{..., $4\times 10^{-1}$, $2\times 10^{-1}$, $1\times 10^{-1}$, $5\times 10^{-2}$, $1\times 10^{-2}$, $5\times 10^{-3}$, ...\}. The server learning rate $\alpha$ is selected from $\{0.2, 0.4, 0.6, 0.8, 0.9, 1.0\}$. The momentum fusion constant $\beta$ is selected from $\{0.2, 0.4, 0.6, 0.8, 0.9, 1.0\}$. Following \cite{karimireddy2020scaffold,hsu2019measuring,wang2020tackling}, we use local epoch $E$ instead of local training steps $P$ in experiments. $E=1$ is identical to one pass training of local data. We test local epoch $E\in\{0.5, 1, 2\}$ and $E=1$ by default.

\textbf{Data Similarity $s$.} We follow \cite{karimireddy2020scaffold} to simulate the \textit{non-i.i.d.} data distribution. Specifically, fraction $s$ of the data are randomly selected and allocated to clients, while the remaining fraction $1-s$ are allocated by sorting according to the label. The data similarity is hence $s$. We run experiments with data similarity $s$ in \{5\%, 10\%, 20\%\}. By default, the data similarity is set to 10\% and the number of clients (GPUs) $K=16$ following \cite{wang2020tackling}. For all experiments, We report the mean and standard deviation metrics in the form of $(mean \pm std)$ over 3 runs with different random seeds for allocating data to clients.

\textbf{Dataset.} We train VGG-16 \cite{simonyan2014very} and ResNet-56 \cite{he2016deep} models on CIFAR-10/100\footnote{https://www.cs.toronto.edu/~kriz/cifar.html} \cite{krizhevsky2009learning}, and ResNet-20 on SVHN\footnote{http://ufldl.stanford.edu/housenumbers/} image classification tasks. Please refer to the Supplementary Material for details.

\subsection{Performance}

We illustrate the experimental results in Figures \ref{fig:training curves vary s}, \ref{fig:heatmap}, and \ref{fig:training curves vary e} with test accuracy ($mean\pm std$) reported in the brackets of the legend, and Table \ref{tab:svhn} and \ref{tab:cifar100}. Testing performance is the main metric for comparison in FL because local training metrics become less meaningful with clients tending to overfit their local data during local training. In overall, \textbf{DOMO(-S) $>$ FedAvgSLM(-Z) $>$ FedAvgSM $>$ FedAvgLM(-Z) $>$ FedAvg} regarding the test accuracy. DOMO and DOMO-S consistently achieve the fastest empirical convergence rate and best test accuracy in all experiments. On the contrary, the initial convergence rate of FedAvgSLM(-Z) can even be worse than FedAvgSM. In particular, FedAvgSLM(-Z) can hurt the performance compared with FedAvgSM as shown in the right plot of Figure \ref{fig:training curves vary e}, possibly due to the theoretical difficulties without momentum fusion discussed in Section ``Convergence of DOMO". Besides, using server statistics is much better than without it (FedAvgSM $\gg$ FedAvg and FedAvgSLM(-Z) $\gg$ FedAvgLM(-Z)), in accordance with \cite{hsu2019measuring,huo2020faster}.

\textbf{Varying Data Similarity $s$.} We plot the training curves under different data similarity settings in Figure \ref{fig:training curves vary s}. We can see that the improvement of DOMO and DOMO-S over other momentum-based methods increases with the data similarity $s$ decreasing. This property makes our proposed method favorable in FL where the data heterogeneity can be complicated. In particular, DOMO improves FedAvgSLM-Z, FedAvgSM, and FedAvg by \textbf{5.00\%, 5.68\%, and 23.14\%} respectively regarding the test accuracy when $s=5\%$. When $s=10\%$ and $s=20\%$, DOMO improves over the best counterpart by \textbf{2.13\%} and \textbf{1.02\%}, while DOMO-S improves by \textbf{1.41\%} and \textbf{0.85\%} respectively.

\textbf{Varying the Server and Local Momentum Constant $\mu_s$, $\mu_l$.} We explore the various combinations of server and local momentum constant $\mu_s$ and $\mu_l$ of DOMO and report the test accuracy in Figure \ref{fig:heatmap}. $\mu_s=0.9$ and $\mu_l=0.6$ work best regardless of the data similarity $s$ and the algorithm we use. Deviating from $\mu_s=0.9$ and $\mu_l=0.6$ leads to gradually lower test accuracy.

\textbf{Varying the Local Epoch $E$.} We plot the training curves of VGG-16 under different local epoch $E$ settings in Figure \ref{fig:training curves vary e} with data similarity $s=10\%$. The number of local training steps $P=49$ and 196 respectively when $E=0.5$ and 2. We can see that DOMO improves the test accuracy over the best counterpart by \textbf{1.25\%} and \textbf{1.70\%} when $E=0.5$ and 2 respectively.

\textbf{Varying Hyper-parameters $\alpha$ and $\beta$.} We explore the combinations of hyper-parameters $\alpha$ and $\beta$ and report the corresponding test accuracy in Table \ref{tab:vary alpha beta}. $\alpha=1.0$ and $\beta=0.9$ work best.

\textbf{Varying Model.} We also plot the training curves of ResNet-56 in the right plot of Figure \ref{fig:training curves vary e} which exhibit a similar pattern. DOMO improves the best counterpart by \textbf{1.35\%} when data similarity $s=10\%$ and local epoch $E=1$. FedAvgSLM(-Z) is inferior to FedAvgSM, implying that a naive combination of FedAvgSM and FedAvgLM can hurt the performance. In contrast, DOMO and DOMO-S improve FedAvgSLM by \textbf{2.45\%} and \textbf{2.03\%} respectively.

\textbf{Varying Dataset.} The SVHN test accuracy is summarized in Table \ref{tab:svhn} and we can see that DOMO and DOMO-S improve the counterpart by \textbf{1.45\%} and \textbf{1.56\%} respectively. The CIFAR-100 test accuracy is summarized in Table \ref{tab:cifar100} and we can see that DOMO and DOMO-S improve the best counterpart by \textbf{2.46\%} and \textbf{1.68\%} respectively when training VGG-16. They improve the counterpart by \textbf{2.39\%} and \textbf{0.89\%} respectively when training ResNet-56.

\section{Conclusion}
In this work, we proposed a new double momentum SGD (DOMO) method with a novel momentum fusion technique to improve the state-of-the-art momentum-based FL algorithm. We provided new insights for the connection between the server and local momentum with new concepts of pre-momentum, intra-momentum, and post-momentum. We also derived the first convergence analysis involving both the server and local Polyak's momentum SGD and discussed the difficulties of theoretical analysis in previous methods that are addressed by DOMO. From a theoretical perspective, we showed that momentum fusion in DOMO could lead to a tighter inconsistency bound. Future works may include incorporating the inter-client variance reduction technique to tighten the divergence bound as well. Deep FL experimental results on benchmark datasets verify the effectiveness of DOMO. DOMO can achieve an improvement of up to 5\% regarding the test accuracy compared with the state-of-the-art momentum-based methods when training VGG-16 on CIFAR-10.

\bibliography{aaai22.bib}

\appendix
\onecolumn
\input{appendix}

\end{document}

%% file: appendix.tex
\section{Task Settings}\label{appendix:task settings}

\textbf{CIFAR-10.} We train VGG-16 and ResNet-56 models on CIFAR-10 image classification task. For VGG-16, there is no batch normalization layer. For ResNet-56, we replace the batch normalization layer with the group normalization layer because \textit{non-i.i.d.} data distribution causes inaccurate batch statistics estimation and worsens the client drift issue. The number of groups in group normalization is set to 8. The local batch size $b=32$ and the total batch size $B=Kb=512$. The weight decay is $5\times 10^{-4}$. The model is trained for 200 epochs with a learning rate decay of 0.1 at epoch 120 and 160. Random cropping, random flipping, and standardization are applied as data augmentation techniques.

\textbf{SVHN.} We train ResNet-20 on SVHN dataset. The number of groups in group normalization is set to 8. This task is simpler than CIFAR-10 and we set $E=5$, $s=2\%$ to enlarge the difference of different methods with a smaller model ResNet-20. The local batch size $b=32$ and the total batch size $B=Kb=512$. The model is trained for 200 epochs with a learning rate decay of 0.1 at epoch 120 and 160. The weight decay is $1\times 10^{-4}$. We do not apply data augmentation in this task.

\textbf{CIFAR-100.} We train VGG-16 and ResNet-56 on CIFAR-100 dataset. The number of groups in group normalization is set to 8. This task is harder than CIFAR-10 and we set $E=0.4$, $s=40\%$ to make all methods converge. The local batch size $b=32$ and the total batch size $B=Kb=512$. The model is trained for 200 epochs with a learning rate decay of 0.1 at epoch 120 and 160. The weight decay is $5\times 10^{-4}$. Random cropping, random flipping, and standardization are applied as data augmentation techniques.

\section{Proof of Theorem 1}\label{appendix:proof}

The update of server momentum $\textbf{m}_r$ and model $\textbf{x}_r$ follows
\begin{equation}
    \textbf{m}_{r+1}=\mu_s \textbf{m}_r+\frac{1}{KP}\sum^{K-1}_{k=0}\sum^{P-1}_{p=0}\textbf{m}^{(k)}_{r,p+1} \quad \text{and} \quad \textbf{x}_{r+1}=\textbf{x}_{r}-\alpha\eta P\textbf{m}_{r+1}\,.
\end{equation}
The update of local momentum $\textbf{m}^{(k)}_{r,p}$ follows momentum SGD excepts that it will be averaged or reset to zero in the end of each training round. To facilitate the analysis involving both the Polyak's server momentum and local momentum for the first time, we propose to define a sequence $\{\textbf{y}_{r}\}$ as
\begin{equation}
\begin{split}
    \textbf{y}_{r} &=\textbf{x}_r + \frac{\mu_s}{1-\mu_s}(\textbf{x}_{r}-\textbf{x}_{r-1}) = \textbf{x}_{r}-\frac{\mu_s}{1-\mu_s}\alpha\eta P\textbf{m}_{r}\\
    &=\frac{1}{1-\mu_s}\textbf{x}_{r}-\frac{\mu_s}{1-\mu_s} \textbf{x}_{r-1} \,\,(r\geq 1) \quad \text{and} \quad \textbf{y}_0=\textbf{x}_0\,.
\end{split}
\end{equation}
We can set $\textbf{x}_{-1}=\textbf{x}_0$ to remove the condition in the bracket. It is easy to see that
\begin{equation}
\begin{split}
    \textbf{y}_{r+1}-\textbf{y}_{r} &= \frac{1}{1-\mu_s}(\textbf{x}_{r+1}-\textbf{x}_{r})-\frac{\mu_s}{1-\mu_s} (\textbf{x}_{r}-\textbf{x}_{r-1})=-\frac{\alpha}{1-\mu_s}\eta P(\textbf{m}_{r+1}-\mu_s \textbf{m}_r)\\
    &= -\frac{\alpha\eta}{(1-\mu_s)K}\sum^{K-1}_{k=0}\sum^{P-1}_{p=0}\textbf{m}^{(k)}_{r,p+1}\,.
\end{split}
\end{equation}
We also let
\begin{equation}
    \widehat{\textbf{y}}_{r,p}=\textbf{y}_r-\frac{\alpha\eta}{(1-\mu_s)K}\sum^{K-1}_{k=0}\sum^{p-1}_{p^\prime=0}\textbf{m}^{(k)}_{r,p^\prime+1} \quad \text{such that} \quad \widehat{\textbf{y}}_{r,0}=\textbf{y}_r \quad \text{and} \quad \widehat{\textbf{y}}_{r,P}=\textbf{y}_{r+1}=\widehat{\textbf{y}}_{r+1,0}\,.
\end{equation}
It is easy to see that
\begin{equation}
    \widehat{\textbf{y}}_{r,p+1}-\widehat{\textbf{y}}_{r,p}=-\frac{\alpha\eta}{(1-\mu_s)K}\sum^{K-1}_{k=0}\textbf{m}^{(k)}_{r,p+1}\,.
\end{equation}
To facilitate the analysis involving local momentum in $\widehat{\textbf{y}}_{r,p}$, we define another novel sequence $\{\textbf{z}_{r,p}\}$ as
\begin{equation}
\begin{split}
    \textbf{z}_{r,p} &=\widehat{\textbf{y}}_{r,p} + \frac{\mu_l}{1-\mu_l}(\widehat{\textbf{y}}_{r,p}-\widehat{\textbf{y}}_{r,p-1}) = \widehat{\textbf{y}}_{r,p}-\frac{\mu_l\alpha\eta}{(1-\mu_l)(1-\mu_s)K}\sum^{K-1}_{k=0}\textbf{m}^{(k)}_{r,p}\\
    &=\frac{1}{1-\mu_l}\widehat{\textbf{y}}_{r,p}-\frac{\mu_l}{1-\mu_l}\widehat{\textbf{y}}_{r,p-1} \,\,(r\geq 1 \,\, \text{or}\,\, p \geq 1) \quad \text{and} \quad \textbf{z}_{0,0}=\widehat{\textbf{y}}_{0,0}=\textbf{y}_0=\textbf{x}_0\,.
\end{split}
\end{equation}
We can set $\widehat{\textbf{y}}_{0,-1}=\widehat{\textbf{y}}_{0,0}$ to remove the condition in the bracket. Then the update of $\textbf{z}_{r,p}$ becomes
\begin{equation}
\begin{split}
    \textbf{z}_{r,p+1}-\textbf{z}_{r,p} &= \frac{1}{1-\mu_l}(\widehat{\textbf{y}}_{r,p+1}-\widehat{\textbf{y}}_{r,p})-\frac{\mu_l}{1-\mu_l}(\widehat{\textbf{y}}_{r,p}-\widehat{\textbf{y}}_{r,p-1})\\
    &=-\frac{\alpha\eta}{(1-\mu_l)(1-\mu_s)K}\sum^{K-1}_{k=0}(\textbf{m}^{(k)}_{r,p+1}-\mu_l \textbf{m}^{(k)}_{r,p}) \,.
\end{split}
\end{equation}
If the local momentum is reset ($\textbf{m}^{(k)}_{r,0}\leftarrow \textbf{0}$) instead of being averaged ($\textbf{m}^{(k)}_{r,0}\leftarrow \frac{1}{K}\sum^{K-1}_{k=0}\textbf{m}^{(k)}_{r-1,P}$) at the end of each training round, we need to separately consider this update rule when $p=-1$ for the last equality in the above equation. Therefore, we consider \textbf{two cases}:

\textbf{(a)} average momentum;

\textbf{(b)} reset momentum. In particular, for this case we have either \textbf{(b.1)} $p\neq-1$ or \textbf{(b.2)} $p=-1$.

For $0\leq p \leq P-1$ (cases (a) and (b.1)), we have
\begin{equation}
    \textbf{z}_{r,p+1}-\textbf{z}_{r,p} = -\frac{\alpha\eta}{(1-\mu_l)(1-\mu_s)K}\sum^{K-1}_{k=0}\nabla F^{(k)}(\textbf{x}^{(k)}_{r,p},\xi^{(k)}_{r,p}) \,.
\end{equation}
We also consider $p=-1$ (case (b.2)) for completeness as $-1$ has been used in some previous definitions:
\begin{equation}
\begin{split}
    &\textbf{z}_{r,p+1}-\textbf{z}_{r,p} = \textbf{z}_{r,0} - \textbf{z}_{r,-1} = \textbf{z}_{r-1,P} - \textbf{z}_{r-1,P-1}\\
    &=-\frac{\alpha\eta}{(1-\mu_l)(1-\mu_s)K}\sum^{K-1}_{k=0}(\textbf{m}^{(k)}_{r-1,P}-\mu_l\textbf{m}_{r-1,P-1}^{(k)})\\
    &= -\frac{\alpha\eta}{(1-\mu_l)(1-\mu_s)K}\sum^{K-1}_{k=0}\nabla F^{(k)}(\textbf{x}^{(k)}_{r-1,P-1},\xi^{(k)}_{r-1,P-1})\,.
\end{split}
\end{equation}
In this way the critical property $\textbf{z}_{r,P}=\textbf{z}_{r+1,0}$ is preserved from $\{\widehat{\textbf{y}}_{r,p}\}$. In contrast, the analysis of the update between $\overline{\textbf{x}}_{r+1,0}$ and $\overline{\textbf{x}}_{r,P-1}$ is more tricky. Now we analyze the convergence of $\{\textbf{z}_{r,p}\}$ and compare its difference with $\{\overline{\textbf{x}}_{r,p}\}$.

\subsection{Inconsistency Bound of $\|\textbf{z}_{r,p}-\overline{\textbf{x}}_{r,p}\|^2_2$ (Lemma 2)}\label{appendix:bound1}
In this section, some procedures in the proof of case (a) and case (b) can be different. Let's first consider case (a).
\begin{equation}
\begin{split}
    \textbf{z}_{r,p} &= \widehat{\textbf{y}}_{r,p} + \frac{\mu_l}{1-\mu_l}(\widehat{\textbf{y}}_{r,p}-\widehat{\textbf{y}}_{r,p-1}) = \widehat{\textbf{y}}_{r,p} - \frac{\mu_l\alpha\eta}{(1-\mu_l)(1-\mu_s)K}\sum^{K-1}_{k=0}\textbf{m}^{(k)}_{r,p}\\
    &=\textbf{y}_{r}-\frac{\alpha\eta}{(1-\mu_s)K}\sum^{K-1}_{k=0}\sum^{p-1}_{p^\prime=0}\textbf{m}^{(k)}_{r,p^\prime+1}-\frac{\mu_l\alpha\eta}{(1-\mu_l)(1-\mu_s)K}\sum^{K-1}_{k=0}\textbf{m}^{(k)}_{r,p}\\
    &=\textbf{x}_{r}-\frac{\mu_s}{1-\mu_s}\alpha\eta P\textbf{m}_{r}-\frac{\alpha\eta}{(1-\mu_s)K}\sum^{K-1}_{k=0}\sum^{p-1}_{p^\prime=0}\textbf{m}^{(k)}_{r,p^\prime+1}-\frac{\mu_l\alpha\eta}{(1-\mu_l)(1-\mu_s)K}\sum^{K-1}_{k=0}\textbf{m}^{(k)}_{r,p}\,.
\end{split}
\end{equation}
For DOMO algorithm, we have
\begin{equation}
\begin{split}
    \overline{\textbf{x}}_{r,p} = \frac{1}{K}\sum^{K-1}_{k=0}\textbf{x}^{(k)}_{r,p}=\textbf{x}_{r} - \beta\eta P\textbf{m}_{r} - \frac{\eta}{K}\sum^{K-1}_{k=0}\sum^{p-1}_{p^\prime=0}\textbf{m}^{(k)}_{r,p^\prime+1}\,.
\end{split}
\end{equation}
Let $\beta=\frac{\mu_s}{1-\mu_s}\alpha$ and we have,
\begin{equation}
    \textbf{z}_{r,p}-\overline{\textbf{x}}_{r,p}=(1-\frac{\alpha}{1-\mu_s})\frac{\eta}{K}\sum^{K-1}_{k=0}\sum^{p-1}_{p^\prime=0}\textbf{m}^{(k)}_{r,p^\prime+1} - \frac{\mu_l\alpha\eta}{(1-\mu_l)(1-\mu_s)K}\sum^{K-1}_{k=0}\textbf{m}^{(k)}_{r,p}\,.
\end{equation}
For ease of notification, we define $t=rP+p$, and $\nabla F^{(k)}_{t}=\nabla F^{(k)}(\textbf{x}^{(k)}_{r,p},\xi^{(k)}_{r,p})$. Then for case (a) we have
\begin{equation}\label{mark1}
\begin{split}
    &\|\textbf{z}_{r,p}-\overline{\textbf{x}}_{r,p}\|^2_2\\
    &= \|\frac{\eta}{K}\sum^{K-1}_{k=0}[(1-\frac{\alpha}{1-\mu_s})\sum^{p-1}_{p^\prime=0}\textbf{m}^{(k)}_{r,p^\prime+1}-\frac{\mu_l}{1-\mu_l}\frac{\alpha}{1-\mu_s}\textbf{m}^{(k)}_{r,p}]\|^2_2\\
    &= \|\frac{\eta}{K}\sum^{K-1}_{k=0}[(1-\frac{\alpha}{1-\mu_s})\sum^{p-1}_{p^\prime=0}\sum^{t-p+p^\prime}_{\tau=0}\mu_l^{t-p+p^\prime-\tau}\nabla F^{(k)}_\tau-\frac{\mu_l}{1-\mu_l}\frac{\alpha}{1-\mu_s}\sum^{t-1}_{\tau=0}\mu_l^{t-1-\tau}\nabla F^{(k)}_\tau]\|^2_2\\
    &= \|\frac{\eta}{K}\sum^{K-1}_{k=0}[(1-\frac{\alpha}{1-\mu_s})\sum^{p-1}_{p^\prime=0}\sum^{t-p}_{\tau=0}\mu_l^{t-p+p^\prime-\tau}\nabla F^{(k)}_\tau-\frac{\mu_l}{1-\mu_l}\frac{\alpha}{1-\mu_s}\sum^{t-p}_{\tau=0}\mu_l^{t-1-\tau}\nabla F^{(k)}_\tau\\
    &\quad +(1-\frac{\alpha}{1-\mu_s})\sum^{p-1}_{p^\prime=0}\sum^{t-p+p^\prime}_{\tau=t-p+1}\mu_l^{t-p+p^\prime-\tau}\nabla F^{(k)}_\tau-\frac{\mu_l}{1-\mu_l}\frac{\alpha}{1-\mu_s}\sum^{t-1}_{\tau=t-p+1}\mu_l^{t-1-\tau}\nabla F^{(k)}_\tau\|^2_2\\
    &= \|\frac{\eta}{K}\sum^{K-1}_{k=0}[(1-\frac{\alpha}{1-\mu_s})\sum^{p-1}_{p^\prime=0}\sum^{t-p}_{\tau=0}\mu_l^{t-p+p^\prime-\tau}\nabla F^{(k)}_\tau-\frac{\mu_l}{1-\mu_l}\frac{\alpha}{1-\mu_s}\sum^{t-p}_{\tau=0}\mu_l^{t-1-\tau}\nabla F^{(k)}_\tau\\
    &\quad +(1-\frac{\alpha}{1-\mu_s})\sum^{t-1}_{\tau=t-p+1}\sum^{p-1}_{p^\prime=\tau-t+p}\mu_l^{t-p+p^\prime-\tau}\nabla F^{(k)}_\tau-\frac{\mu_l}{1-\mu_l}\frac{\alpha}{1-\mu_s}\sum^{t-1}_{\tau=t-p+1}\mu_l^{t-1-\tau}\nabla F^{(k)}_\tau\|^2_2\\
    &= \|\frac{\eta}{K}\sum^{K-1}_{k=0}[\sum^{t-p}_{\tau=0}\mu_l^{t-1-\tau}\nabla F^{(k)}_\tau((1-\frac{\alpha}{1-\mu_s})\sum^{p-1}_{p^\prime=0}\mu_l^{-p+p^\prime+1}-\frac{\mu_l}{1-\mu_l}\frac{\alpha}{1-\mu_s})\\
    &\quad +\sum^{t-1}_{\tau=t-p+1}\mu_l^{t-1-\tau}\nabla F^{(k)}_\tau((1-\frac{\alpha}{1-\mu_s})\sum^{p-1}_{p^\prime=\tau-t+p}\mu_l^{-p+p^\prime+1}-\frac{\mu_l}{1-\mu_l}\frac{\alpha}{1-\mu_s})]\|^2_2 \,.\\
\end{split}
\end{equation}
For simplicity, let
\begin{equation}
\begin{split}
    & h_1=\frac{\mu_l}{1-\mu_l}\frac{\alpha}{1-\mu_s}-(1-\frac{\alpha}{1-\mu_s})\sum^{p-1}_{p^\prime=0}\mu_l^{-p+p^\prime+1}=\frac{\alpha}{1-\mu_s}\frac{1+\mu_l-\mu_l^p}{1-\mu_l} - \frac{1-\mu_l^p}{1-\mu_l}\\
    & \leq h \coloneqq \frac{\alpha}{1-\mu_s}\frac{1+\mu_l-\mu_l^p}{1-\mu_l} - 1
\end{split}
\end{equation}
\begin{equation}
    h_2=\frac{\mu_l}{1-\mu_l}\frac{\alpha}{1-\mu_s}-(1-\frac{\alpha}{1-\mu_s})\sum^{p-1}_{p^\prime=\tau-t+p}\mu_l^{-p+p^\prime+1}=\frac{\alpha}{1-\mu_s}\frac{1+\mu_l-\mu_l^{t-\tau}}{1-\mu_l} - \frac{1-\mu_l^{t-\tau}}{1-\mu_l}
\end{equation}
When $t-p+1\leq \tau \leq t-1$, i.e., $1 \leq t-\tau \leq p-1 < p$, we have $h_2 < h$. Suppose
\begin{equation}
    \alpha \geq (1-\mu_s)(1-\mu_l) \geq \max\{\frac{(1-\mu_s)(1-\mu_l^p)}{1+\mu_l-\mu_l^p}, \frac{(1-\mu_s)(1-\mu_l^{t-\tau})}{1+\mu_l-\mu_l^{t-\tau}}\}
\end{equation}
Then it is easy to see that $h_1,h_2 \geq 0$.

\begin{equation}\label{appendix:mark1}
\begin{split}
    &\|\textbf{z}_{r,p}-\overline{\textbf{x}}_{r,p}\|^2_2\\
    &\leq \|\frac{\eta}{K}\sum^{K-1}_{k=0}(\sum^{t-p}_{\tau=0}h_1\mu_l^{t-1-\tau}\nabla F^{(k)}_\tau+\sum^{t-1}_{\tau=t-p+1}h_2\mu_l^{t-1-\tau}\nabla F^{(k)}_\tau)\|^2_2\\
    &\leq \eta^2(\sum^{t-p}_{\tau=0}h_1\mu_l^{t-1-\tau}+\sum^{t-1}_{\tau=t-p+1}h_2\mu_l^{t-1-\tau})\\
    &\quad\cdot(\sum^{t-p}_{\tau=0}h_1\mu_l^{t-1-\tau}\|\frac{1}{K}\sum^{K-1}_{k=0}\nabla F^{(k)}_\tau\|^2_2 + \sum^{t-1}_{\tau=t-p+1}h_2\mu_l^{t-1-\tau}\|\frac{1}{K}\sum^{K-1}_{k=0}\nabla F^{(k)}_\tau\|^2_2)\\
    &\leq \eta^2(\sum^{t-1}_{\tau=0}h\mu_l^{t-1-\tau})\sum^{t-1}_{\tau=0}h\mu_l^{t-1-\tau}\|\frac{1}{K}\sum^{K-1}_{k=0}\nabla F^{(k)}_\tau\|^2_2\\
    &\leq \frac{\eta^2h^2}{1-\mu_l}\sum^{t-1}_{\tau=0}\mu_l^{t-1-\tau}\|\frac{1}{K}\sum^{K-1}_{k=0}\nabla F^{(k)}_\tau\|^2_2\,.
\end{split}
\end{equation}
Let $T=RP$. Note that $h$ is a function of $\alpha$ and $p$ (or $t$). Summing from $t=0$ to $T-1$ yields
\begin{equation}
\begin{split}
    &\sum^{T-1}_{t=0}\|\textbf{z}_{r,p}-\overline{\textbf{x}}_{r,p}\|^2_2 \leq \frac{\eta^2}{1-\mu_l}\sum^{T-1}_{t=0}h^2\sum^{t-1}_{\tau=0}\mu_l^{t-1-\tau}\|\frac{1}{K}\sum^{K-1}_{k=0}\nabla F^{(k)}_\tau\|^2_2\\
    &=\frac{\eta^2}{1-\mu_l}\sum^{T-2}_{\tau=0}\|\frac{1}{K}\sum^{K-1}_{k=0}\nabla F^{(k)}_\tau\|^2_2\sum^{T-1}_{t=\tau+1}h^2\mu_l^{t-1-\tau}\\
    &\leq \frac{\eta^2}{1-\mu_l}\sum^{T-2}_{\tau=0}\|\frac{1}{K}\sum^{K-1}_{k=0}\nabla F^{(k)}_\tau\|^2_2\sum^{+\infty}_{t=0}h^2\mu_l^{t}\\
    &\leq \frac{\eta^2}{1-\mu_l}\sum^{T-2}_{\tau=0}\|\frac{1}{K}\sum^{K-1}_{k=0}\nabla F^{(k)}_\tau\|^2_2\sum^{P-1}_{p=0}h
    ^2\sum^{+\infty}_{n=0}\mu_l^{p+nP}\\
    &= \frac{\eta^2}{1-\mu_l}\sum^{T-2}_{\tau=0}\|\frac{1}{K}\sum^{K-1}_{k=0}\nabla F^{(k)}_\tau\|^2_2\sum^{P-1}_{p=0}h^2\frac{\mu_l^p}{1-\mu_l^P}\,.\\
\end{split}
\end{equation}
The R.H.S. is minimized when $\alpha=(1-\mu_s)(1-\mu_l)$, $h1-=\mu_l-\mu_l^p$ and we have
\begin{equation}
\begin{split}
    &\sum^{T-1}_{t=0}\|\textbf{z}_{r,p}-\overline{\textbf{x}}_{r,p}\|^2_2 \leq \frac{\eta^2}{1-\mu_l}\sum^{T-2}_{\tau=0}\|\frac{1}{K}\sum^{K-1}_{k=0}\nabla F^{(k)}_\tau\|^2_2\sum^{P-1}_{p=0}\mu_l^2\frac{\mu_l^p}{1-\mu_l^P} = \frac{\eta^2\mu_l^2}{(1-\mu_l)^2}\sum^{T-1}_{t=0}\|\frac{1}{K}\sum^{K-1}_{k=0}\nabla F^{(k)}_t\|^2_2\,.
\end{split}
\end{equation}
In particular, for local momentum SGD, $\alpha=1$, $\mu_s=0$, and $h_1=h_2=\frac{\mu_l}{1-\mu_l}$. Moreover, for DOMO with $\alpha=1-\mu_s$ and $\beta=\mu_s$, we still have $h_1=h_2=\frac{\mu_l}{1-\mu_l}$. For both of them, we can get the inconsistency bound following the above precedure by replacing $h$ with $\frac{\mu_l}{1-\mu_l}$:
\begin{equation}
    \sum^{T-1}_{t=0}\|\textbf{z}_{r,p}-\overline{\textbf{x}}_{r,p}\|^2_2 \leq \frac{\eta^2\mu_l^2}{(1-\mu_l)^4}\sum^{T-1}_{t=0}\|\frac{1}{K}\sum^{K-1}_{k=0}\nabla F^{(k)}_t\|^2_2\,.
\end{equation}
Compare the above two upper bounds and we can see that DOMO achieves $(1-\mu_l)^2$ of the inconsistency bound in local momentum SGD. However, we note that a potential higher improvement may be achieved as $\alpha=(1-\mu_s)(1-\mu_l)$ may not be optimal due to inequality scaling. Therefore in experiments hyper-parameters $\alpha$ and $\beta$ require further tuning.

As the improvement is a constant factor and does not affect the convergence rate (though it helps empirical training), in later analysis we simply set $\alpha=1$ and $\beta=\mu_s$ to preserve the same inconsistency bound.

Following similar procedures except that the local momentum is reset to $\textbf{0}$ every $P$ iterations, for case (b) we have
\begin{equation}
\begin{split}
    &\|\textbf{z}_{r,p}-\overline{\textbf{x}}_{r,p}\|^2_2 = \|\frac{\eta}{K}\sum^{K-1}_{k=0}[(1-\frac{\alpha}{1-\mu_s})\sum^{p-1}_{p^\prime=0}\textbf{m}^{(k)}_{r,p^\prime+1}-\frac{\mu_l}{1-\mu_l}\textbf{m}^{(k)}_{r,p}]\|^2_2\\
    &= \|\frac{\eta}{K}\sum^{K-1}_{k=0}[(1-\frac{\alpha}{1-\mu_s})\sum^{p-1}_{p^\prime=0}\sum^{t-p+p^\prime}_{\tau=t-p}\mu_l^{t-p+p^\prime-\tau}\nabla F^{(k)}_\tau-\frac{\mu_l}{1-\mu_l}\sum^{t-1}_{\tau=t-p}\mu_l^{t-1-\tau}\nabla F^{(k)}_\tau]\|^2_2\\
    &= \|\frac{\eta}{K}\sum^{K-1}_{k=0}[(1-\frac{\alpha}{1-\mu_s})\sum^{t-1}_{\tau=t-p}\sum^{p-1}_{p^\prime=\tau-t+p}\mu_l^{t-p+p^\prime-\tau}\nabla F^{(k)}_\tau-\frac{\mu_l}{1-\mu_l}\sum^{t-1}_{\tau=t-p}\mu_l^{t-1-\tau}\nabla F^{(k)}_\tau]\|^2_2\\
    &= \|\frac{\eta}{K}\sum^{K-1}_{k=0}\sum^{t-1}_{\tau=t-p}\mu_l^{t-1-\tau}\nabla F_{\tau}^{(k)}h_2\|^2_2 \leq \|\eta h_1\sum^{t-1}_{\tau=t-p}\mu_l^{t-1-\tau}\sum^{K-1}_{k=0}\frac{1}{K}\nabla F_{\tau}^{(k)}\|^2_2\\
    &\leq \frac{\eta^2h_1^2}{1-\mu_l}\sum^{t-1}_{\tau=t-p}\mu_l^{t-1-\tau}\|\frac{1}{K}\sum^{K-1}_{k=0}\nabla F^{(k)}_\tau\|^2_2 \leq \frac{\eta^2h_1^2}{1-\mu_l}\sum^{t-1}_{\tau=0}\mu_l^{t-1-\tau}\|\frac{1}{K}\sum^{K-1}_{k=0}\nabla F^{(k)}_\tau\|^2_2 \,.
\end{split}
\end{equation}
Compare it with Eq. (\ref{appendix:mark1}) and we can see that the inconsistency bound in case (a) can also bound that in case (b).

\subsection{Divergence Bound of $\|\overline{\textbf{x}}_{r,p}-\textbf{x}^{(k)}_{r,p}\|^2_2$}\label{appendix:bound2}
We note that in this section, the proof is identical for either case (a) or case (b). We first consider
\begin{equation}\label{diff1}
\begin{split}
    &\frac{1}{K}\sum^{K-1}_{k=0}\|\nabla f^{(k)}(\textbf{x}^{(k)}_{r,p})-\frac{1}{K}\sum^{K-1}_{k^\prime=0}\nabla f^{(k^\prime)}(\textbf{x}^{(k^\prime)}_{r,p})\|^2_2 \\
    &\leq \frac{1}{K}\sum^{K-1}_{k=0}(3\|\nabla f^{(k)}(\textbf{x}^{(k)}_{r,p})-\nabla f^{(k)}(\overline{\textbf{x}}_{r,p})\|^2_2 + 3\|\nabla f^{(k)}(\overline{\textbf{x}}_{r,p})-\nabla f(\overline{\textbf{x}}_{r,p})\|^2_2\\
    &\quad + 3\|\nabla f(\overline{\textbf{x}}_{r,p})-\frac{1}{K}\sum^{K-1}_{k^\prime=0}\nabla f^{(k^\prime)}(\textbf{x}^{(k^\prime)}_{r,p})\|^2_2)\\
    &\leq \frac{6L^2}{K}\sum^{K-1}_{k=0}\|\textbf{x}^{(k)}_{r,p}-\overline{\textbf{x}}_{r,p}\|^2_2 + 3G^2\,.
\end{split}
\end{equation}
Then,
\begin{equation}
\begin{split}
    &\frac{1}{K}\sum^{K-1}_{k=0}\mathbb{E}\|\overline{\textbf{x}}_{r,p}-\textbf{x}^{(k)}_{r,p}\|^2_2\\
    &= \frac{1}{K}\sum^{K-1}_{k=0}\mathbb{E}\|\frac{1}{K}\sum^{K-1}_{k^\prime=0}(\textbf{x}_{r}-\mu_s\eta P\textbf{m}_r-\sum^{p-1}_{p^\prime=0}\eta\textbf{m}^{(k^\prime)}_{r,p^\prime+1}) - (\textbf{x}_{r}-\mu_s\eta P\textbf{m}_r-\sum^{p-1}_{p^\prime=0}\eta\textbf{m}^{(k)}_{r,p^\prime+1})\|^2_2\\
    &=\frac{\eta^2}{K}\sum^{K-1}_{k=0}\mathbb{E}\|\sum^{p-1}_{p^\prime=0}(\frac{1}{K}\sum^{K-1}_{k^\prime=0}\textbf{m}^{(k^\prime)}_{r,p^\prime+1}-\textbf{m}^{(k)}_{r,p^\prime+1})\|^2_2\\
    &=\frac{\eta^2}{K}\sum^{K-1}_{k=0}\mathbb{E}\|\sum^{p-1}_{p^\prime=0}(\frac{1}{K}\sum^{K-1}_{k^\prime=0}\nabla F^{(k^\prime)}(\textbf{x}^{(k^\prime)}_{r,p^\prime},\xi^{(k^\prime)}_{r,p^\prime})-\nabla F^{(k)}(\textbf{x}^{(k)}_{r,p^\prime},\xi^{(k)}_{r,p^\prime}))\frac{1-\mu_l^{p-p^\prime}}{1-\mu_l}\|^2_2\\
    &\leq \frac{2\eta^2}{K}\sum^{K-1}_{k=0}\mathbb{E}\|\sum^{p-1}_{p^\prime=0}[\frac{1}{K}\sum^{K-1}_{k^\prime=0}(\nabla F^{(k^\prime)}(\textbf{x}^{(k^\prime)}_{r,p^\prime},\xi^{(k^\prime)}_{r,p^\prime})-\nabla f^{(k)}(\textbf{x}^{(k^\prime)}_{r,p^\prime}))\\
    &\quad\quad\quad\quad\quad\quad\quad\quad\quad-(\nabla F^{(k)}(\textbf{x}^{(k)}_{r,p^\prime},\xi^{(k)}_{r,p^\prime})-\nabla f^{(k)}(\textbf{x}^{(k)}_{r,p^\prime}))]\frac{1-\mu_l^{p-p^\prime}}{1-\mu_l}\|^2_2\\
    &\quad +\frac{2\eta^2}{K}\sum^{K-1}_{k=0}\mathbb{E}\|\sum^{p-1}_{p^\prime=0}[\frac{1}{K}\sum^{K-1}_{k^\prime=0}\nabla f^{(k^\prime)}(\textbf{x}^{(k^\prime)}_{r,p^\prime})-\nabla f^{(k)}(\textbf{x}^{(k)}_{r,p^\prime})]\frac{1-\mu_l^{p-p^\prime}}{1-\mu_l}\|^2_2\\
    &\leq \frac{2\eta^2P\sigma^2}{(1-\mu_l)^2} + \frac{2\eta^2P}{(1-\mu_l)^2K}\sum^{K-1}_{k=0}\sum^{p-1}_{p^\prime=0}\mathbb{E}\|\frac{1}{K}\sum^{K-1}_{k^\prime=0}\nabla f^{(k^\prime)}(\textbf{x}^{(k^\prime)}_{r,p^\prime})-\nabla f^{(k)}(\textbf{x}^{(k)}_{r,p^\prime})\|^2_2\\
    &\leq \frac{2\eta^2P\sigma^2}{(1-\mu_l)^2} + \frac{2\eta^2P}{(1-\mu_l)^2}\sum^{p-1}_{p^\prime=0}(\frac{6L^2}{K}\sum^{K-1}_{k=0}\mathbb{E}\|\overline{\textbf{x}}_{r,p^\prime}-\textbf{x}^{(k)}_{r,p^\prime}\|^2_2+3G^2)\\
    &\leq \frac{12\eta^2PL^2}{(1-\mu_l)^2K}\sum^{p-1}_{p^\prime=0}\sum^{K-1}_{k=0}\mathbb{E}\|\overline{\textbf{x}}_{r,p^\prime}-\textbf{x}^{(k)}_{r,p^\prime}\|^2_2 + \frac{2\eta^2P\sigma^2}{(1-\mu_l)^2} + \frac{6\eta^2P^2G^2}{(1-\mu_l)^2}
\end{split}
\end{equation}
Sum $t$ from $0$ to $T-1=RP-1$ (i.e., sum $p$ from $0$ to $P-1$ and sum $r$ from $0$ to $R-1$) and let $P\leq \frac{1-\mu_l}{6\eta L}$,
\begin{equation}
\begin{split}
    &\frac{1}{KT}\sum^{T-1}_{t=0}\sum^{K-1}_{k=0}\mathbb{E}\|\overline{\textbf{x}}_{r,p}-\textbf{x}^{(k)}_{r,p}\|^2_2\\
    &\leq \frac{12\eta^2P^2L^2}{(1-\mu_l)^2KT}\sum^{T-1}_{t=0}\sum^{K-1}_{k=0}\mathbb{E}\|\overline{\textbf{x}}_{r,p}-\textbf{x}^{(k)}_{r,p}\|^2_2 + \frac{2\eta^2P\sigma^2}{(1-\mu_l)^2} + \frac{6\eta^2P^2G^2}{(1-\mu_l)^2}\\
    &\leq \frac{1}{3KT}\sum^{T-1}_{t=0}\sum^{K-1}_{k=0}\mathbb{E}\|\overline{\textbf{x}}_{r,p}-\textbf{x}^{(k)}_{r,p}\|^2_2 + \frac{2\eta^2P\sigma^2}{(1-\mu_l)^2} + \frac{6\eta^2P^2G^2}{(1-\mu_l)^2}\\
    &\leq \frac{3\eta^2P\sigma^2}{(1-\mu_l)^2} + \frac{9\eta^2P^2G^2}{(1-\mu_l)^2}\,.
\end{split}
\end{equation}

\subsection{Main Proof}
Consider the improvement in one training round ($0\leq p\leq P-1$). By the smoothness assumption,
\begin{equation}
\begin{split}
    &\mathbb{E}_{r,p}f(\textbf{z}_{r,p+1})-\mathbb{E}_{r,p}f(\textbf{z}_{r,p}) \leq \mathbb{E}_{r,p}\langle\nabla f(\textbf{z}_{r,p}),\textbf{z}_{r,p+1}-\textbf{z}_{r,p}\rangle + \frac{L}{2}\mathbb{E}_{r,p}\|\textbf{z}_{r,p+1}-\textbf{z}_{r,p}\|^2_2\\
    &= -\frac{\eta}{1-\mu_l}\langle\nabla f(\textbf{z}_{r,p}),\frac{1}{K}\sum^{K-1}_{k=0}\nabla f^{(k)}(\textbf{x}^{(k)}_{r,p})\rangle + \frac{L\eta^2}{2(1-\mu_l)^2}\mathbb{E}_{r,p}\|\frac{1}{K}\sum^{K-1}_{k=0}\nabla F^{(k)}(\textbf{x}^{(k)}_{r,p},\xi^{(k)}_{r,p})\|^2_2\\
    &= -\frac{\eta}{1-\mu_l}\langle\nabla f(\textbf{z}_{r,p}),\frac{1}{K}\sum^{K-1}_{k=0}\nabla f^{(k)}(\textbf{x}^{(k)}_{r,p})\rangle + \frac{L\eta^2}{2(1-\mu_l)^2}\mathbb{E}_{r,p}\|\frac{1}{K}\sum^{K-1}_{k=0}\nabla f^{(k)}(\textbf{x}^{(k)}_{r,p})\|^2_2\\
    &\quad + \frac{L\eta^2\sigma^2}{2(1-\mu_l)^2K}\,.
\end{split}
\end{equation}
$\forall \gamma\in\mathbb{R}^+$, the first term
\begin{equation}\label{diff2}
\begin{split}
    &-\frac{\eta}{1-\mu_l}\langle\nabla f(\textbf{z}_{r,p}),\frac{1}{K}\sum^{K-1}_{k=0}\nabla f^{(k)}(\textbf{x}^{(k)}_{r,p})\rangle\\
    &= -\frac{\eta}{1-\mu_l}\langle\nabla f(\textbf{z}_{r,p})-\nabla f(\overline{\textbf{x}}_{r,p}),\frac{1}{K}\sum^{K-1}_{k=0}\nabla f^{(k)}(\textbf{x}^{(k)}_{r,p})\rangle\\
    &\quad - \frac{\eta}{1-\mu_l}\langle\nabla f(\overline{\textbf{x}}_{r,p}),\frac{1}{K}\sum^{K-1}_{k=0}\nabla f^{(k)}(\textbf{x}^{(k)}_{r,p})\rangle\\
    &\leq \frac{\eta}{2(1-\mu_l)\gamma}\|\nabla f(\textbf{z}_{r,p})-\nabla f(\overline{\textbf{x}}_{r,p})\|^2_2 + \frac{\eta\gamma}{2(1-\mu_l)}\|\frac{1}{K}\sum^{K-1}_{k=0}\nabla f^{(k)}(\textbf{x}^{(k)}_{r,p})\|^2_2\\
    & \quad -\frac{\eta}{2(1-\mu_l)}(\|\nabla f(\overline{\textbf{x}}_{r,p})\|^2_2 + \|\frac{1}{K}\sum^{K-1}_{k=0}\nabla f^{(k)}(\textbf{x}^{(k)}_{r,p})\|^2_2-\|\nabla f(\overline{\textbf{x}}_{r,p})-\frac{1}{K}\sum^{K-1}_{k=0}\nabla f^{(k)}(\textbf{x}^{(k)}_{r,p})\|^2_2)\\
    &\leq \frac{\eta L^2}{2(1-\mu_l)\gamma}\|\textbf{z}_{r,p}-\overline{\textbf{x}}_{r,p}\|^2_2 - \frac{\eta(1-\gamma)}{2(1-\mu_l)}\|\frac{1}{K}\sum^{K-1}_{k=0}\nabla f^{(k)}(\textbf{x}^{(k)}_{r,p})\|^2_2 - \frac{\eta}{2(1-\mu_l)}\|\nabla f(\overline{\textbf{x}}_{r,p})\|^2_2\\
    &\quad + \frac{\eta L^2}{2(1-\mu_l)K}\sum^{K-1}_{k=0}\|\overline{\textbf{x}}_{r,p}-\textbf{x}^{(k)}_{r,p}\|^2_2\,.
\end{split}
\end{equation}
Combine the above two equations,
\begin{equation}
\begin{split}
    &\mathbb{E}_{r,p}f(\textbf{z}_{r,p+1})-\mathbb{E}_{r,p}f(\textbf{z}_{r,p})\\
    &\leq - \frac{\eta}{2(1-\mu_l)}\|\nabla f(\overline{\textbf{x}}_{r,p})\|^2_2 + \frac{\eta L^2}{2(1-\mu_l)\gamma}\|\textbf{z}_{r,p}-\overline{\textbf{x}}_{r,p}\|^2_2 + \frac{\eta L^2}{2(1-\mu_l)K}\sum^{K-1}_{k=0}\|\overline{\textbf{x}}_{r,p}-\textbf{x}^{(k)}_{r,p}\|^2_2\\
    &\quad + \frac{L\eta^2\sigma^2}{2(1-\mu_l)^2K} - \frac{\eta}{2(1-\mu_l)}(1-\gamma-L\eta)\|\frac{1}{K}\sum^{K-1}_{k=0}\nabla f^{(k)}(\textbf{x}^{(k)}_{r,p})\|^2_2\,.
\end{split}
\end{equation}
Now we take the total expectation and sum $t$ from $0$ to $T-1=RP-1$ (i.e., sum $p$ from $0$ to $P-1$ and sum $r$ from $0$ to $R-1$),
\begin{equation}
\begin{split}
    &\frac{1}{T}[\mathbb{E}f(\textbf{z}_{R-1,P})-f(\textbf{z}_{0,0})] \leq -\frac{\eta}{2(1-\mu_l)T}\sum^{T-1}_{t=0}\mathbb{E}\|\nabla f(\overline{\textbf{x}}_{r,p})\|^2_2 + \frac{\eta L^2}{2\gamma(1-\mu_l)T}\sum^{T-1}_{t=0}\mathbb{E}\|\textbf{z}_{r,p}-\overline{\textbf{x}}_{r,p}\|^2_2\\
    &\quad\quad\quad\quad\quad\quad\quad\quad\quad\quad\quad+ \frac{\eta L^2}{2(1-\mu_l)TK}\sum^{T-1}_{t=0}\sum^{K-1}_{k=0}\mathbb{E}\|\overline{\textbf{x}}_{r,p}-\textbf{x}^{(k)}_{r,p}\|^2_2 + \frac{L\eta^2\sigma^2}{2(1-\mu_l)^2K}\\
    &\quad\quad\quad\quad\quad\quad\quad\quad\quad\quad\quad- \frac{\eta}{2(1-\mu_l)T}(1-\gamma-L\eta)\sum^{T-1}_{t=0}\mathbb{E}\|\frac{1}{K}\sum^{K-1}_{k=0}\nabla f^{(k)}(\textbf{x}^{(k)}_{r,p})\|^2_2\,.
\end{split}
\end{equation}
Based on the bound derived in section \ref{appendix:bound1} and \ref{appendix:bound2}, let $\gamma=\frac{1}{2}$ and $1-\gamma-L\eta-\frac{\mu_l^2\eta^2L^2}{\gamma(1-\mu_l)^4}\geq 0$, we have
\begin{equation}
\begin{split}
    &\frac{1}{T}[f_*-f(\textbf{x}_{0})] \leq \frac{1}{T}[\mathbb{E}f(\textbf{z}_{R-1,P})-f(\textbf{z}_{0,0})]\\
    &\leq -\frac{\eta}{2(1-\mu_l)T}\sum^{T-1}_{t=0}\mathbb{E}\|\nabla f(\overline{\textbf{x}}_{r,p})\|^2_2 + \frac{\eta L^2}{2\gamma(1-\mu_l)T}\frac{\mu_l^2\eta^2}{(1-\mu_l)^4}\sum^{T-1}_{t=0}\|\frac{1}{K}\sum^{K-1}_{k=0}\nabla F(\textbf{x}^{(k)}_{r,p},\xi^{(k)}_{r,p})\|^2_2\\
    &\quad + \frac{\eta L^2}{2(1-\mu_l)}[\frac{3\eta^2P\sigma^2}{(1-\mu_l)^2} + \frac{9\eta^2P^2G^2}{(1-\mu_l)^2}]\\
    &\quad  + \frac{L\eta^2\sigma^2}{2(1-\mu_l)^2K} - \frac{\eta}{2(1-\mu_l)T}(1-\gamma-L\eta)\sum^{T-1}_{t=0}\mathbb{E}\|\frac{1}{K}\sum^{K-1}_{k=0}\nabla f^{(k)}(\textbf{x}^{(k)}_{r,p})\|^2_2\\
    &= -\frac{\eta}{2(1-\mu_l)T}\sum^{T-1}_{t=0}\mathbb{E}\|\nabla f(\overline{\textbf{x}}_{r,p})\|^2_2 + \frac{\eta^2L\sigma^2}{2(1-\mu_l)^2}(\frac{1}{K}+\frac{3\eta LP}{2(1-\mu_l)}+\frac{\mu_l^2\eta L}{\gamma(1-\mu_l)^4K})\\
    &\quad + \frac{9\eta^3L^2P^2G^2}{2(1-\mu_l)^3} - \frac{\eta}{2(1-\mu_l)T}(1-\gamma-L\eta-\frac{\mu_l^2\eta^2L^2}{\gamma(1-\mu_l)^4})\sum^{T-1}_{t=0}\mathbb{E}\|\frac{1}{K}\sum^{K-1}_{k=0}\nabla f^{(k)}(\textbf{x}^{(k)}_{r,p})\|^2_2\\
    &\leq -\frac{\eta}{2(1-\mu_l)T}\sum^{T-1}_{t=0}\mathbb{E}\|\nabla f(\overline{\textbf{x}}_{r,p})\|^2_2 + \frac{\eta^2L\sigma^2}{2(1-\mu_l)^2}(\frac{1}{K}+\frac{3\eta LP}{2(1-\mu_l)}+\frac{2\mu_l^2\eta L}{(1-\mu_l)^4K})\\
    &\quad + \frac{9\eta^3L^2P^2G^2}{2(1-\mu_l)^3}\,.\\
\end{split}
\end{equation}
Rearrange and we have
\begin{equation}
\begin{split}
    \frac{1}{T}\sum^{T-1}_{t=0}\mathbb{E}\|\nabla f(\overline{\textbf{x}}_{r,p})\|^2_2 &\leq \frac{2(1-\mu_l)(f(\textbf{x}_0)-f_*)}{\eta T} + \frac{\eta L\sigma^2}{(1-\mu_l)}(\frac{1}{K}+\frac{3\eta LP}{2(1-\mu_l)}+\frac{2\mu_l^2\eta L}{(1-\mu_l)^4K})\\
    &\quad +\frac{9\eta^2L^2P^2G^2}{(1-\mu_l)^2}\,,
\end{split}
\end{equation}
which completes the proof.

\section{Extension to Partial Participation}

In this section we show that it is possible to extend the theoretical analysis of DOMO to cross-device FL with partial clients participation in each training round.

For the algorithm side, suppose we randomly sample a client set $\mathcal{V}_r$ to participate in training round $r$. Let $|\mathcal{V}_r|=S$. For the computation of client $k$, we can just replace client $k\in[K]$ (full participation) with $k\in \mathcal{V}\_r$ (partial participation). Correspondingly, the server should only communicate with clients in $\mathcal{V}_r$.

That is, $k\in[K]\rightarrow k\in\mathcal{V}_r$ and $\frac{1}{K}\sum^{K-1}_{k=0} \rightarrow \frac{1}{S}\sum_{k\in\mathcal{V}_r}$.

For the convergence analysis, we should also do such replacement. But besides the replacement, we note that now $\nabla f(\textbf{x})=\frac{1}{K}\sum^{K-1}_{k=0}\nabla f^{(k)}(\textbf{x}) \neq \frac{1}{S}\sum_{k\in\mathcal{V}_r}\nabla f^{(k)}(\textbf{x})$, which will affect the following two inequalities of the proof in the previous section.

(a) Eq. (\ref{diff1}):
\begin{equation}
\begin{split}
    \frac{1}{K}\sum^{K-1}_{k=0}\|\nabla f^{(k)}(\textbf{x}^{(k)}_{r,p})-\frac{1}{K}\sum^{K-1}_{k^\prime=0}\nabla f^{(k^\prime)}(\textbf{x}^{(k^\prime)}_{r,p})\|^2_2  \leq  \frac{6L^2}{K}\sum^{K-1}_{k=0}\|\textbf{x}^{(k)}_{r,p}-\overline{\textbf{x}}_{r,p}\|^2_2 + 3G^2
\end{split}
\end{equation}

New (partial participation):
\begin{equation}
\begin{split}
    &\frac{1}{S}\sum_{k\in\mathcal{V}_r}\|\nabla f^{(k)}(\textbf{x}^{(k)}_{r,p})-\frac{1}{S}\sum_{k^\prime\in\mathcal{V}_r}\nabla f^{(k^\prime)}(\textbf{x}^{(k^\prime)}_{r,p})\|^2_2\\
    &\leq \frac{1}{S}\sum_{k\in\mathcal{V}_r}\|\nabla f^{(k)}(\textbf{x}^{(k)}_{r,p})-\frac{1}{S}\sum_{k^\prime\in\mathcal{V}_r}\nabla f^{(k^\prime)}(\textbf{x}^{(k^\prime)}_{r,p})\|^2_2\\
    & \leq \frac{1}{S}\sum_{k\in\mathcal{V}_r}[3\|\nabla f^{(k)}(\textbf{x}^{(k)}_{r,p})-\nabla f^{(k)}(\overline{\textbf{x}}_{r,p})\|^2_2    +    3\|\frac{1}{S}\sum_{k^\prime\in\mathcal{V}_r}(\nabla f^{(k^\prime)}(\overline{\textbf{x}}_{r,p}) - \nabla f^{(k^\prime)}(\textbf{x}^{(k^\prime)}_{r,p}) )\|^2_2\\
    &    +    3\|\nabla f^{(k)}(\overline{\textbf{x}}_{r,p}) - \frac{1}{S}\sum_{k\in\mathcal{V}_r}\nabla f^{(k^\prime)}(\overline{\textbf{x}}_{r,p})\|^2_2]\\
    &\leq  \frac{6L^2}{S}\sum_{k\in\mathcal{V}_r}\|\textbf{x}^{(k)}_{r,p}-\overline{\textbf{x}}_{r,p}\|^2_2 + \frac{1}{S}\sum_{k\in\mathcal{V}_r}[6\|\nabla f^{(k)}(\overline{\textbf{x}}_{r,p}) - \nabla f(\overline{\textbf{x}}_{r,p})\|^2_2 + 6\|\nabla f(\overline{\textbf{x}}_{r,p}) - \frac{1}{S}\sum_{k^\prime\in\mathcal{V}_r}\nabla f^{(k^\prime)}(\overline{\textbf{x}}_{r,p})\|^2_2]\\
    &\leq  \frac{6L^2}{S}\sum_{k\in\mathcal{V}_r}\|\textbf{x}^{(k)}_{r,p}-\overline{\textbf{x}}_{r,p}\|^2_2 + 12G^2
\end{split}
\end{equation}

This larger constant coefficient of $G^2$ does not affect the convergence rate. This leads to a new bound in section \ref{appendix:bound2}
\begin{equation}
    \frac{1}{KT}\sum^{T-1}_{t=0}\sum^{K-1}_{k=0}\mathbb{E}\|\overline{\textbf{x}}_{r,p}-\textbf{x}_{r,p}^{(k)}\|^2_2 \leq \frac{3\eta^2P^2\sigma^2}{(1-\mu_l)^2} + \frac{36\eta^2P^2G^2}{(1-\mu_l)^2}
\end{equation}

(b) In Eq. (\ref{diff2}), we showed that
\begin{equation}
\begin{split}
    &- \frac{\eta}{1-\mu_l}\langle\nabla f(\overline{\textbf{x}}_{r,p}),\frac{1}{K}\sum^{K-1}_{k=0}\nabla f^{(k)}(\textbf{x}^{(k)}_{r,p})\rangle\\
    &\leq - \frac{\eta}{2(1-\mu_l)}\|\nabla f(\overline{\textbf{x}}_{r,p})\|^2_2       -            \frac{\eta}{2(1-\mu_l)}\|\frac{1}{K}\sum^{K-1}_{k=0}\nabla f^{(k)}(\textbf{x}^{(k)}_{r,p})\|^2_2       +       \frac{\eta L^2}{2(1-\mu_l)K}\sum^{K-1}_{k=0}\|\overline{\textbf{x}}_{r,p}-\textbf{x}^{(k)}_{r,p}\|^2_2
\end{split}
\end{equation}

New (partial participation): suppose the client set $\mathcal{V}_r$ is randomly and uniformly sampled following common practice, such that $\mathbb{E}_{\mathcal{V}_r}[\frac{1}{S}\sum_{k\in\mathcal{V}_r}\nabla f^{(k)}(\textbf{x})]=\nabla f(\textbf{x})$. Then
\begin{equation}
\begin{split}
    &- \frac{\eta}{1-\mu_l}\mathbb{E}_{\mathcal{V}_r}\langle\nabla f(\overline{\textbf{x}}_{r,p}),\frac{1}{S}\sum_{k\in\mathcal{V}_r}\nabla f^{(k)}(\textbf{x}^{(k)}_{r,p})\rangle\\
    &= - \frac{\eta}{1-\mu_l}\mathbb{E}_{\mathcal{V}_r}\langle\nabla f(\overline{\textbf{x}}_{r,p}),\frac{1}{S}\sum_{k\in\mathcal{V}_r}\nabla f^{(k)}(\overline{\textbf{x}}_{r,p})\rangle   -    \frac{\eta}{1-\mu_l}\mathbb{E}_{\mathcal{V}_r}\langle\nabla f(\overline{\textbf{x}}_{r,p}),\frac{1}{S}\sum_{k\in\mathcal{V}_r}(\nabla f^{(k)}(\textbf{x}^{(k)}_{r,p})-\nabla f^{(k)}(\overline{\textbf{x}}_{r,p}))\\
    &\leq - \frac{\eta}{1-\mu_l}\|\nabla f(\overline{\textbf{x}}_{r,p})\|^2_2 + \frac{\eta}{2(1-\mu_l)}\mathbb{E}_{\mathcal{V}_r}[\|\nabla f(\overline{\textbf{x}}_{r,p})\|^2_2 + \|\frac{1}{S}\sum_{k\in\mathcal{V}_r}(\nabla f^{(k)}(\textbf{x}^{(k)}_{r,p})-\nabla f^{(k)}(\overline{\textbf{x}}_{r,p}))\|^2_2]\\
    &\leq - \frac{\eta}{2(1-\mu_l)}\|\nabla f(\overline{\textbf{x}}_{r,p})\|^2_2  +  \mathbb{E}_{\mathcal{V}_r}[\frac{\eta L^2}{2(1-\mu_l)S}\sum_{k\in\mathcal{V}_r}\|\textbf{x}^{(k)}_{r,p}-\overline{\textbf{x}}_{r,p}\|^2_2]
\end{split}
\end{equation}

Compared with the previous result, this bound is larger by the term
$$\frac{\eta}{2(1-\mu_l)}\|\frac{1}{S}\sum_{k\in\mathcal{V}_r}\nabla f^{(k)}(\textbf{x}^{(k)}_{r,p})\|^2_2\,,$$
which we will need to bound in the main proof. A simple way to bound it will be an additional assumption to bound the gradient norm. In summary, not much of the current proof in the previous section needs to be modified. For partial participation, we will need a a stronger assumption that $\|\nabla f^{(k)}(\textbf{x})\|^2_2\leq M^2$. Following the modification above and the main proof, we will have a convergence rate $\mathcal{O}(S^{\frac{1}{2}}R^{-\frac{1}{2}}P^{-\frac{1}{2}})$.

% \begin{equation}
% \begin{split}
%     &\frac{1}{T}[f_*-f(\textbf{x}_{0})]\\
%     &\leq -\frac{\eta}{2(1-\mu_l)T}\sum^{T-1}_{t=0}\mathbb{E}\|\nabla f(\overline{\textbf{x}}_{r,p})\|^2_2 + \frac{\eta^2L\sigma^2}{2(1-\mu_l)^2}(\frac{1}{K}+\frac{3\eta LP}{2(1-\mu_l)}+\frac{\mu_l^2\eta L}{\gamma(1-\mu_l)^4K})\\
%     &\quad + \frac{36\eta^3L^2P^2G^2}{2(1-\mu_l)^3} + \frac{\eta}{2(1-\mu_l)T}(\gamma+L\eta+\frac{\mu_l^2\eta^2L^2}{\gamma(1-\mu_l)^4})\sum^{T-1}_{t=0}\mathbb{E}\|\frac{1}{K}\sum^{K-1}_{k=0}\nabla f^{(k)}(\textbf{x}^{(k)}_{r,p})\|^2_2\\
%     &\leq -\frac{\eta}{2(1-\mu_l)T}\sum^{T-1}_{t=0}\mathbb{E}\|\nabla f(\overline{\textbf{x}}_{r,p})\|^2_2 + \frac{\eta^2L\sigma^2}{2(1-\mu_l)^2}(\frac{1}{K}+\frac{3\eta LP}{2(1-\mu_l)}+\frac{2\mu_l^2\eta L}{(1-\mu_l)^4K})\\
%     &\quad + \frac{36\eta^3L^2P^2G^2}{2(1-\mu_l)^3} + \frac{\eta}{2(1-\mu_l)}(\frac{1}{2}+L\eta+\frac{2\mu_l^2\eta^2L^2}{(1-\mu_l)^4})M^2\,.\\
% \end{split}
% \end{equation}
% Rearrange and we have
% \begin{equation}
% \begin{split}
%     \frac{1}{T}\sum^{T-1}_{t=0}\mathbb{E}\|\nabla f(\overline{\textbf{x}}_{r,p})\|^2_2 &\leq \frac{2(1-\mu_l)(f(\textbf{x}_0)-f_*)}{\eta T} + \frac{\eta L\sigma^2}{(1-\mu_l)}(\frac{1}{K}+\frac{3\eta LP}{2(1-\mu_l)}+\frac{2\mu_l^2\eta L}{(1-\mu_l)^4K})\\
%     &\quad +\frac{36\eta^2L^2P^2G^2}{(1-\mu_l)^2} + (\frac{1}{2}+L\eta+\frac{2\mu_l^2\eta^2L^2}{(1-\mu_l)^4})M^2\,,
% \end{split}
% \end{equation}

%% file: cameral_ready.bbl
\begin{thebibliography}{50}
\providecommand{\natexlab}[1]{#1}

\bibitem[{Acar et~al.(2020)Acar, Zhao, Matas, Mattina, Whatmough, and
  Saligrama}]{acar2020federated}
Acar, D. A.~E.; Zhao, Y.; Matas, R.; Mattina, M.; Whatmough, P.; and Saligrama,
  V. 2020.
\newblock Federated learning based on dynamic regularization.
\newblock In \emph{International Conference on Learning Representations}.

\bibitem[{Bonawitz et~al.(2019)Bonawitz, Eichner, Grieskamp, Huba, Ingerman,
  Ivanov, Kiddon, Kone{\v{c}}n{\`y}, Mazzocchi, McMahan
  et~al.}]{bonawitz2019towards}
Bonawitz, K.; Eichner, H.; Grieskamp, W.; Huba, D.; Ingerman, A.; Ivanov, V.;
  Kiddon, C.; Kone{\v{c}}n{\`y}, J.; Mazzocchi, S.; McMahan, H.~B.; et~al.
  2019.
\newblock Towards federated learning at scale: System design.
\newblock \emph{arXiv preprint arXiv:1902.01046}.

\bibitem[{Chen and Huo(2016)}]{chen2016scalable}
Chen, K.; and Huo, Q. 2016.
\newblock Scalable training of deep learning machines by incremental block
  training with intra-block parallel optimization and blockwise model-update
  filtering.
\newblock In \emph{2016 ieee international conference on acoustics, speech and
  signal processing (icassp)}, 5880--5884. IEEE.

\bibitem[{Cutkosky and Orabona(2019)}]{cutkosky2019momentum}
Cutkosky, A.; and Orabona, F. 2019.
\newblock Momentum-based variance reduction in non-convex sgd.
\newblock In \emph{Advances in Neural Information Processing Systems},
  15236--15245.

\bibitem[{Defazio, Bach, and Lacoste-Julien(2014)}]{defazio2014saga}
Defazio, A.; Bach, F.; and Lacoste-Julien, S. 2014.
\newblock SAGA: A fast incremental gradient method with support for
  non-strongly convex composite objectives.
\newblock \emph{Advances in neural information processing systems}, 27:
  1646--1654.

\bibitem[{Defazio and Bottou(2018)}]{defazio2018ineffectiveness}
Defazio, A.; and Bottou, L. 2018.
\newblock On the ineffectiveness of variance reduced optimization for deep
  learning.
\newblock \emph{arXiv preprint arXiv:1812.04529}.

\bibitem[{Duchi, Hazan, and Singer(2011)}]{duchi2011adaptive}
Duchi, J.; Hazan, E.; and Singer, Y. 2011.
\newblock Adaptive subgradient methods for online learning and stochastic
  optimization.
\newblock \emph{Journal of machine learning research}, 12(7).

\bibitem[{Fallah, Mokhtari, and Ozdaglar(2020)}]{fallah2020personalized}
Fallah, A.; Mokhtari, A.; and Ozdaglar, A. 2020.
\newblock Personalized Federated Learning with Theoretical Guarantees: A
  Model-Agnostic Meta-Learning Approach.
\newblock \emph{Advances in Neural Information Processing Systems}, 33.

\bibitem[{Gao, Xu, and Huang(2021)}]{gao2021convergence}
Gao, H.; Xu, A.; and Huang, H. 2021.
\newblock On the Convergence of Communication-Efficient Local SGD for Federated
  Learning.
\newblock In \emph{Proceedings of the AAAI Conference on Artificial
  Intelligence}, volume~35.

\bibitem[{Gu et~al.(2021)Gu, Xu, Huo, Deng, and Huang}]{gu2021privacy}
Gu, B.; Xu, A.; Huo, Z.; Deng, C.; and Huang, H. 2021.
\newblock Privacy-Preserving Asynchronous Vertical Federated Learning
  Algorithms for Multiparty Collaborative Learning.
\newblock \emph{IEEE Transactions on Neural Networks and Learning Systems}.

\bibitem[{Guo et~al.(2021)Guo, Wang, Zhou, Jiang, and Patel}]{guo2021multi}
Guo, P.; Wang, P.; Zhou, J.; Jiang, S.; and Patel, V.~M. 2021.
\newblock Multi-institutional collaborations for improving deep learning-based
  magnetic resonance image reconstruction using federated learning.
\newblock In \emph{Proceedings of the IEEE/CVF Conference on Computer Vision
  and Pattern Recognition}, 2423--2432.

\bibitem[{He et~al.(2016)He, Zhang, Ren, and Sun}]{he2016deep}
He, K.; Zhang, X.; Ren, S.; and Sun, J. 2016.
\newblock Deep residual learning for image recognition.
\newblock In \emph{Proceedings of the IEEE conference on computer vision and
  pattern recognition}, 770--778.

\bibitem[{Hsieh et~al.(2020)Hsieh, Phanishayee, Mutlu, and
  Gibbons}]{hsieh2020non}
Hsieh, K.; Phanishayee, A.; Mutlu, O.; and Gibbons, P. 2020.
\newblock The non-iid data quagmire of decentralized machine learning.
\newblock In \emph{International Conference on Machine Learning}, 4387--4398.
  PMLR.

\bibitem[{Hsu, Qi, and Brown(2019)}]{hsu2019measuring}
Hsu, T.-M.~H.; Qi, H.; and Brown, M. 2019.
\newblock Measuring the effects of non-identical data distribution for
  federated visual classification.
\newblock \emph{arXiv preprint arXiv:1909.06335}.

\bibitem[{Huo et~al.(2020)Huo, Yang, Gu, Huang et~al.}]{huo2020faster}
Huo, Z.; Yang, Q.; Gu, B.; Huang, L.~C.; et~al. 2020.
\newblock Faster on-device training using new federated momentum algorithm.
\newblock \emph{arXiv preprint arXiv:2002.02090}.

\bibitem[{Ioffe and Szegedy(2015)}]{ioffe2015batch}
Ioffe, S.; and Szegedy, C. 2015.
\newblock Batch normalization: Accelerating deep network training by reducing
  internal covariate shift.
\newblock In \emph{International conference on machine learning}, 448--456.
  PMLR.

\bibitem[{Jiang et~al.(2019)Jiang, Kone{\v{c}}n{\`y}, Rush, and
  Kannan}]{jiang2019improving}
Jiang, Y.; Kone{\v{c}}n{\`y}, J.; Rush, K.; and Kannan, S. 2019.
\newblock Improving federated learning personalization via model agnostic meta
  learning.
\newblock \emph{arXiv preprint arXiv:1909.12488}.

\bibitem[{Johnson and Zhang(2013)}]{johnson2013accelerating}
Johnson, R.; and Zhang, T. 2013.
\newblock Accelerating stochastic gradient descent using predictive variance
  reduction.
\newblock \emph{Advances in neural information processing systems}, 26:
  315--323.

\bibitem[{Kairouz et~al.(2019)Kairouz, McMahan, Avent, Bellet, Bennis, Bhagoji,
  Bonawitz, Charles, Cormode, Cummings et~al.}]{kairouz2019advances}
Kairouz, P.; McMahan, H.~B.; Avent, B.; Bellet, A.; Bennis, M.; Bhagoji, A.~N.;
  Bonawitz, K.; Charles, Z.; Cormode, G.; Cummings, R.; et~al. 2019.
\newblock Advances and open problems in federated learning.
\newblock \emph{arXiv preprint arXiv:1912.04977}.

\bibitem[{Karimireddy et~al.(2020{\natexlab{a}})Karimireddy, Jaggi, Kale,
  Mohri, Reddi, Stich, and Suresh}]{karimireddy2020mime}
Karimireddy, S.~P.; Jaggi, M.; Kale, S.; Mohri, M.; Reddi, S.~J.; Stich, S.~U.;
  and Suresh, A.~T. 2020{\natexlab{a}}.
\newblock Mime: Mimicking centralized stochastic algorithms in federated
  learning.
\newblock \emph{arXiv preprint arXiv:2008.03606}.

\bibitem[{Karimireddy et~al.(2020{\natexlab{b}})Karimireddy, Kale, Mohri,
  Reddi, Stich, and Suresh}]{karimireddy2020scaffold}
Karimireddy, S.~P.; Kale, S.; Mohri, M.; Reddi, S.; Stich, S.; and Suresh,
  A.~T. 2020{\natexlab{b}}.
\newblock Scaffold: Stochastic controlled averaging for federated learning.
\newblock In \emph{International Conference on Machine Learning}, 5132--5143.
  PMLR.

\bibitem[{Karimireddy et~al.(2019)Karimireddy, Rebjock, Stich, and
  Jaggi}]{karimireddy2019error}
Karimireddy, S.~P.; Rebjock, Q.; Stich, S.; and Jaggi, M. 2019.
\newblock Error feedback fixes signsgd and other gradient compression schemes.
\newblock In \emph{International Conference on Machine Learning}, 3252--3261.
  PMLR.

\bibitem[{Kingma and Ba(2014)}]{kingma2014adam}
Kingma, D.~P.; and Ba, J. 2014.
\newblock Adam: A method for stochastic optimization.
\newblock \emph{arXiv preprint arXiv:1412.6980}.

\bibitem[{Kone{\v{c}}n{\`y} et~al.(2016)Kone{\v{c}}n{\`y}, McMahan, Yu,
  Richt{\'a}rik, Suresh, and Bacon}]{konevcny2016federated}
Kone{\v{c}}n{\`y}, J.; McMahan, H.~B.; Yu, F.~X.; Richt{\'a}rik, P.; Suresh,
  A.~T.; and Bacon, D. 2016.
\newblock Federated learning: Strategies for improving communication
  efficiency.
\newblock \emph{arXiv preprint arXiv:1610.05492}.

\bibitem[{Krizhevsky, Hinton et~al.(2009)}]{krizhevsky2009learning}
Krizhevsky, A.; Hinton, G.; et~al. 2009.
\newblock Learning multiple layers of features from tiny images.

\bibitem[{Li et~al.(2021)Li, Hu, Beirami, and Smith}]{li2021ditto}
Li, T.; Hu, S.; Beirami, A.; and Smith, V. 2021.
\newblock Ditto: Fair and robust federated learning through personalization.
\newblock In \emph{International Conference on Machine Learning}, 6357--6368.
  PMLR.

\bibitem[{Li et~al.(2018)Li, Sahu, Zaheer, Sanjabi, Talwalkar, and
  Smith}]{li2018federated}
Li, T.; Sahu, A.~K.; Zaheer, M.; Sanjabi, M.; Talwalkar, A.; and Smith, V.
  2018.
\newblock Federated optimization in heterogeneous networks.
\newblock \emph{arXiv preprint arXiv:1812.06127}.

\bibitem[{Li et~al.(2020)Li, Sanjabi, Beirami, and Smith}]{Li2020Fair}
Li, T.; Sanjabi, M.; Beirami, A.; and Smith, V. 2020.
\newblock Fair Resource Allocation in Federated Learning.
\newblock In \emph{International Conference on Learning Representations}.

\bibitem[{Liang et~al.(2019)Liang, Shen, Liu, Pan, Chen, and
  Cheng}]{liang2019variance}
Liang, X.; Shen, S.; Liu, J.; Pan, Z.; Chen, E.; and Cheng, Y. 2019.
\newblock Variance reduced local SGD with lower communication complexity.
\newblock \emph{arXiv preprint arXiv:1912.12844}.

\bibitem[{Liu et~al.(2021)Liu, Chen, Qin, Dou, and Heng}]{liu2021feddg}
Liu, Q.; Chen, C.; Qin, J.; Dou, Q.; and Heng, P.-A. 2021.
\newblock Feddg: Federated domain generalization on medical image segmentation
  via episodic learning in continuous frequency space.
\newblock In \emph{Proceedings of the IEEE/CVF Conference on Computer Vision
  and Pattern Recognition}, 1013--1023.

\bibitem[{McMahan et~al.(2017)McMahan, Moore, Ramage, Hampson, and
  y~Arcas}]{mcmahan2017communication}
McMahan, B.; Moore, E.; Ramage, D.; Hampson, S.; and y~Arcas, B.~A. 2017.
\newblock Communication-efficient learning of deep networks from decentralized
  data.
\newblock In \emph{Artificial Intelligence and Statistics}, 1273--1282. PMLR.

\bibitem[{Mohri, Sivek, and Suresh(2019)}]{mohri2019agnostic}
Mohri, M.; Sivek, G.; and Suresh, A.~T. 2019.
\newblock Agnostic Federated Learning.
\newblock In \emph{International Conference on Machine Learning}, 4615--4625.

\bibitem[{Paszke et~al.(2019)Paszke, Gross, Massa, Lerer, Bradbury, Chanan,
  Killeen, Lin, Gimelshein, Antiga et~al.}]{paszke2019pytorch}
Paszke, A.; Gross, S.; Massa, F.; Lerer, A.; Bradbury, J.; Chanan, G.; Killeen,
  T.; Lin, Z.; Gimelshein, N.; Antiga, L.; et~al. 2019.
\newblock PyTorch: An Imperative Style, High-Performance Deep Learning Library.
\newblock \emph{Advances in Neural Information Processing Systems}, 32:
  8026--8037.

\bibitem[{Reddi et~al.(2020)Reddi, Charles, Zaheer, Garrett, Rush,
  Kone{\v{c}}n{\`y}, Kumar, and McMahan}]{reddi2020adaptive}
Reddi, S.; Charles, Z.; Zaheer, M.; Garrett, Z.; Rush, K.; Kone{\v{c}}n{\`y},
  J.; Kumar, S.; and McMahan, H.~B. 2020.
\newblock Adaptive Federated Optimization.
\newblock \emph{arXiv preprint arXiv:2003.00295}.

\bibitem[{Rothchild et~al.(2020)Rothchild, Panda, Ullah, Ivkin, Stoica,
  Braverman, Gonzalez, and Arora}]{rothchild2020fetchsgd}
Rothchild, D.; Panda, A.; Ullah, E.; Ivkin, N.; Stoica, I.; Braverman, V.;
  Gonzalez, J.; and Arora, R. 2020.
\newblock Fetchsgd: Communication-efficient federated learning with sketching.
\newblock In \emph{International Conference on Machine Learning}, 8253--8265.
  PMLR.

\bibitem[{Seide and Agarwal(2016)}]{seide2016cntk}
Seide, F.; and Agarwal, A. 2016.
\newblock CNTK: Microsoft's open-source deep-learning toolkit.
\newblock In \emph{Proceedings of the 22nd ACM SIGKDD International Conference
  on Knowledge Discovery and Data Mining}, 2135--2135.

\bibitem[{Shamsian et~al.(2021)Shamsian, Navon, Fetaya, and
  Chechik}]{shamsian2021personalized}
Shamsian, A.; Navon, A.; Fetaya, E.; and Chechik, G. 2021.
\newblock Personalized Federated Learning using Hypernetworks.
\newblock \emph{arXiv preprint arXiv:2103.04628}.

\bibitem[{Simonyan and Zisserman(2014)}]{simonyan2014very}
Simonyan, K.; and Zisserman, A. 2014.
\newblock Very deep convolutional networks for large-scale image recognition.
\newblock \emph{arXiv preprint arXiv:1409.1556}.

\bibitem[{Srivastava et~al.(2014)Srivastava, Hinton, Krizhevsky, Sutskever, and
  Salakhutdinov}]{srivastava2014dropout}
Srivastava, N.; Hinton, G.; Krizhevsky, A.; Sutskever, I.; and Salakhutdinov,
  R. 2014.
\newblock Dropout: a simple way to prevent neural networks from overfitting.
\newblock \emph{The journal of machine learning research}, 15(1): 1929--1958.

\bibitem[{Stich(2018)}]{stich2018local}
Stich, S.~U. 2018.
\newblock Local SGD Converges Fast and Communicates Little.
\newblock In \emph{International Conference on Learning Representations}.

\bibitem[{Sutskever et~al.(2013)Sutskever, Martens, Dahl, and
  Hinton}]{sutskever2013importance}
Sutskever, I.; Martens, J.; Dahl, G.; and Hinton, G. 2013.
\newblock On the importance of initialization and momentum in deep learning.
\newblock In \emph{International conference on machine learning}, 1139--1147.

\bibitem[{T~Dinh, Tran, and Nguyen(2020)}]{t2020personalized}
T~Dinh, C.; Tran, N.; and Nguyen, T.~D. 2020.
\newblock Personalized Federated Learning with Moreau Envelopes.
\newblock \emph{Advances in Neural Information Processing Systems}, 33.

\bibitem[{Wang and Joshi(2018)}]{wang2018adaptive}
Wang, J.; and Joshi, G. 2018.
\newblock Adaptive communication strategies to achieve the best error-runtime
  trade-off in local-update SGD.
\newblock \emph{arXiv preprint arXiv:1810.08313}.

\bibitem[{Wang et~al.(2020)Wang, Liu, Liang, Joshi, and
  Poor}]{wang2020tackling}
Wang, J.; Liu, Q.; Liang, H.; Joshi, G.; and Poor, H.~V. 2020.
\newblock Tackling the Objective Inconsistency Problem in Heterogeneous
  Federated Optimization.
\newblock In \emph{Advances in Neural Information Processing Systems},
  volume~33, 7611--7623.

\bibitem[{Wang et~al.(2019)Wang, Tantia, Ballas, and Rabbat}]{wang2019slowmo}
Wang, J.; Tantia, V.; Ballas, N.; and Rabbat, M. 2019.
\newblock SlowMo: Improving Communication-Efficient Distributed SGD with Slow
  Momentum.
\newblock In \emph{International Conference on Learning Representations}.

\bibitem[{Xu, Huo, and Huang(2020{\natexlab{a}})}]{xu2020acceleration}
Xu, A.; Huo, Z.; and Huang, H. 2020{\natexlab{a}}.
\newblock On the acceleration of deep learning model parallelism with
  staleness.
\newblock In \emph{Proceedings of the IEEE/CVF Conference on Computer Vision
  and Pattern Recognition}, 2088--2097.

\bibitem[{Xu, Huo, and Huang(2020{\natexlab{b}})}]{xu2020optimal}
Xu, A.; Huo, Z.; and Huang, H. 2020{\natexlab{b}}.
\newblock Optimal gradient quantization condition for communication-efficient
  distributed training.
\newblock \emph{arXiv preprint arXiv:2002.11082}.

\bibitem[{Xu, Huo, and Huang(2021)}]{xu2021step}
Xu, A.; Huo, Z.; and Huang, H. 2021.
\newblock Step-Ahead Error Feedback for Distributed Training with Compressed
  Gradient.

\bibitem[{Yu, Jin, and Yang(2019)}]{yu2019linear}
Yu, H.; Jin, R.; and Yang, S. 2019.
\newblock On the Linear Speedup Analysis of Communication Efficient Momentum
  SGD for Distributed Non-Convex Optimization.
\newblock In \emph{International Conference on Machine Learning}, 7184--7193.

\bibitem[{Zaheer et~al.(2018)Zaheer, Reddi, Sachan, Kale, and
  Kumar}]{zaheer2018adaptive}
Zaheer, M.; Reddi, S.; Sachan, D.; Kale, S.; and Kumar, S. 2018.
\newblock Adaptive methods for nonconvex optimization.
\newblock In \emph{Advances in neural information processing systems},
  9793--9803.

\end{thebibliography}
